\documentclass{mindlab}

\usepackage[toc,header]{appendix}
\usepackage{calc}
\usepackage{float}
\usepackage{pifont}
\usepackage{makecell}
\usepackage{enumitem}
\usepackage{mathrsfs}
\usepackage{adjustbox}
\usepackage{multicol}
\usepackage{changepage}
\usepackage{wrapfig}
\usepackage{amssymb}
\usepackage{array}
\usepackage{colortbl}
\usepackage{tikz}
\usepackage{svg}

\newcolumntype{C}[1]{>{\centering\arraybackslash}p{#1}}
\newcolumntype{M}[1]{>{\centering\arraybackslash}m{#1}}
\newcommand{\apphead}{\rowcolor{mindlabbluepale!45}}
\newcommand{\appgroup}[2]{\addlinespace[2pt]\rowcolor{mindlabbluepale!45}\multicolumn{#1}{@{}l}{\sffamily\bfseries\color{mindlabfg}#2}\\}
\newcommand{\appkey}[1]{{\sffamily\bfseries\color{mindlabblue}#1}}

\newcommand{\githubrow}[2]{\url{https://github.com/#1} & {\color{mindlabmuted}#2}\\[-1pt]}
\newcommand{\githubrepos}{%
  \metadata[\faGithub\ Code]{%
    \begin{tabular}[t]{@{}ll@{}}
      \githubrow{MindLab-Research/mindlab-toolkit}{client SDK}
      \githubrow{verl-project/verl-mint}{open-source server with verl}
      \githubrow{MindLab-Research/mint-cookbook}{cookbook, practices, AutoResearch}
    \end{tabular}%
  }%
}

\newcommand{\fittowidth}[1]{%
  \sbox0{#1}%
  \ifdim\wd0>\textwidth
    \resizebox{\textwidth}{!}{\usebox0}%
  \else
    \usebox0%
  \fi
}

\title{MinT: Managed Infrastructure for Training\\ and Serving Millions of LLMs}
\author{Mind Lab}

\correspondence{\email{contact@mindlab.ltd}}
\githubrepos
\date{May 2026}

\abstract{
\beginabstract
We present MindLab Toolkit (MinT), a managed infrastructure system for Low-Rank Adaptation (LoRA) post-training and online serving. MinT targets a setting where many trained policies are produced over a small number of expensive base-model deployments. Instead of materializing each policy as a merged full checkpoint, MinT keeps the base model resident and moves exported LoRA adapter revisions through rollout, update, export, evaluation, serving, and rollback. This adapter-revision path hides distributed training, serving, scheduling, and data movement behind a service interface, making large-scale LoRA RL easier to run and reproduce.

MinT scales this adapter-revision path along three axes. \textbf{Scale Up} extends LoRA RL to frontier-scale dense and Mixture-of-Experts (MoE) architectures, including MLA and DSA attention paths via LoRA target mapping and rollout correction, while model-parallel training and serving paths are validated beyond 1T total parameters. \textbf{Scale Down} minimizes the training-serving handoff by moving only the exported LoRA adapter, which can be less than 1\% of the base-model size in compact rank-1 settings, eliminating full-checkpoint materialization entirely. In our measurements, adapter-only handoff reduces the measured handoff step by $18.3\times$ on a 4B dense model and by $2.85\times$ on a 30B MoE model; under the same resident-base allocation, concurrent multi-policy Group Relative Policy Optimization (GRPO) shortens wall time by $1.77\times$ and $1.45\times$, respectively, without increasing peak memory. \textbf{Scale Out} expands the policy namespace while keeping engine-local execution bounded. MinT's multi-serve layer exposes a million-scale adapter namespace, validated on a $10^6$-adapter packed LoRA catalog, while each engine keeps bounded CPU-cache and GPU-batch working sets for only the selected revisions it serves. To reduce cold first-touch cost, MinT treats adapter loading as scheduled service work, uses packed MoE LoRA tensors to improve live engine loading by $8.5$--$8.7\times$, and exposes newly registered adapters only after readiness. Together, MinT makes million-scale LoRA policy catalogs operational while training and serving selected adapter revisions over shared 1T-class base models.
}

\begin{document}
\maketitle

\section{Introduction}

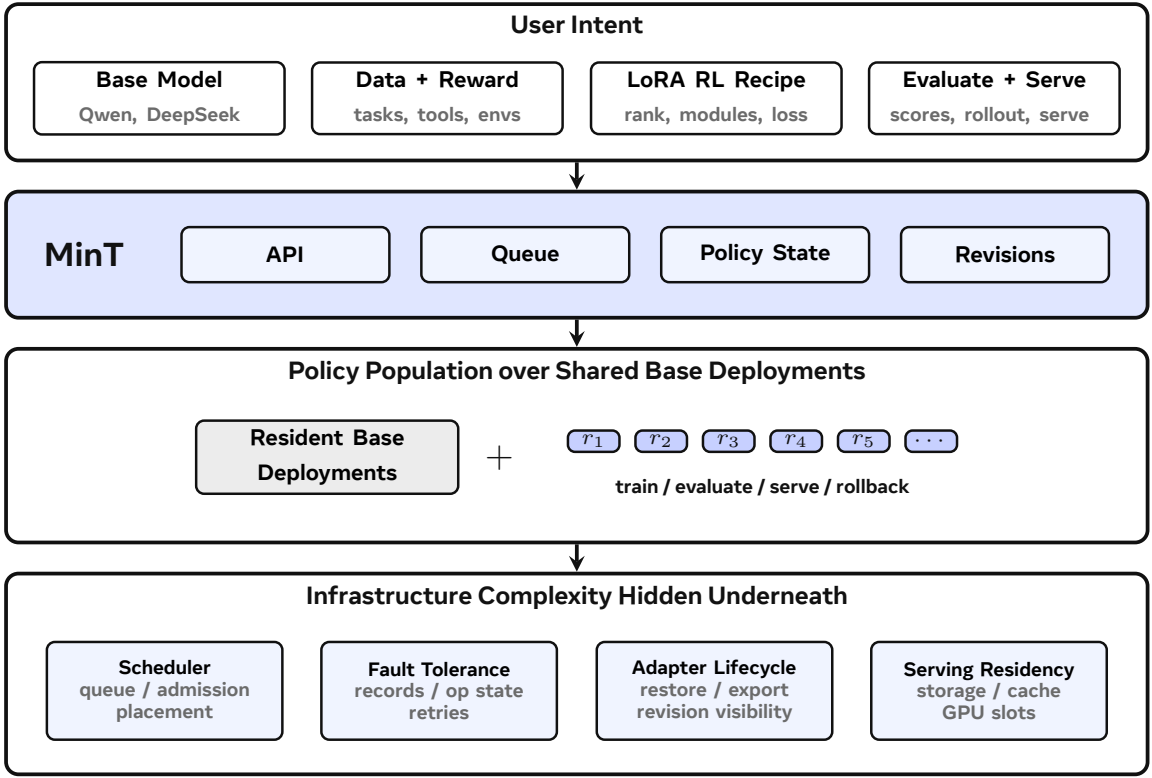
\begin{figure}[t!]
    \centering
    \resizebox{\textwidth}{!}{\begin{tikzpicture}[x=1cm,y=1cm,>=stealth,font=\sffamily]
  \tikzstyle{panel}=[draw=mindlabfg, very thick, rounded corners=1.4mm, fill=white]
  \tikzstyle{mintpanel}=[draw=mindlabfg, very thick, rounded corners=1.4mm, fill=mindlabblue!12]

  \tikzstyle{box}=[
    draw=mindlabfg,
    thick,
    rounded corners=0.9mm,
    fill=white,
    align=center,
    inner sep=3pt
  ]

  \tikzstyle{softbox}=[
    box,
    fill=mindlabbluepale!45,
    minimum width=2.34cm,
    minimum height=0.62cm,
    text width=2.10cm
  ]

  \tikzstyle{groupbox}=[
    draw=mindlabfg,
    thick,
    rounded corners=0.9mm,
    fill=mindlabbluepale!45,
    align=center,
    inner sep=3pt,
    font=\scriptsize\sffamily
  ]

  \tikzstyle{adapter}=[
    draw=mindlabfg,
    thick,
    rounded corners=0.8mm,
    fill=mindlabblue!22,
    minimum width=0.58cm,
    minimum height=0.24cm,
    inner sep=0.6pt,
    font=\scriptsize\sffamily,
    text=mindlabfg
  ]

  \tikzstyle{base}=[
    draw=mindlabfg,
    thick,
    rounded corners=0.9mm,
    fill=mindlabfg!8,
    align=center,
    inner sep=3pt,
    text width=2.55cm
  ]

  \tikzstyle{title}=[
    font=\bfseries\small\sffamily,
    align=center,
    text=mindlabfg
  ]

  \tikzstyle{bigtag}=[
    font=\bfseries\large\sffamily,
    align=center,
    text=mindlabfg
  ]

  \tikzstyle{label}=[
    font=\scriptsize\sffamily,
    align=center,
    text=mindlabfg,
    fill=white,
    inner sep=1pt
  ]

  \tikzstyle{plainlabel}=[
    font=\scriptsize\sffamily,
    align=center,
    text=mindlabfg,
    inner sep=1pt
  ]

  \tikzstyle{arrow}=[->, very thick, draw=mindlabfg]

  \path[use as bounding box] (0.10,-1.02) rectangle (14.10,7.95);

  \node[panel, minimum width=12.85cm, minimum height=1.76cm] (user) at (7.10,7.00) {};
  \node[title] at (7.10,7.65) {User Intent};

  \foreach \x/\main/\sub in {
    2.40/Base Model/{Qwen, DeepSeek},
    5.53/Data + Reward/{tasks, tools, envs},
    8.67/LoRA RL Recipe/{rank, modules, loss},
    11.80/Evaluate + Serve/{scores, rollout, serve}
  } {
    \node[
      box,
      minimum width=2.82cm,
      minimum height=0.62cm,
      text width=2.50cm
    ] at (\x,6.82)
      {{\bfseries\footnotesize \main}\\[-1pt]
       {\scriptsize\color{mindlabfg!62}\sub}};
  }

  \node[mintpanel, minimum width=12.85cm, minimum height=1.42cm] (mint) at (7.10,5.06) {};
  \node[bigtag] at (1.55,5.06) {MinT};

  \foreach \x/\txt in {
    3.82/API,
    6.52/Queue,
    9.22/Policy State,
    11.92/Revisions
  } {
    \node[softbox] at (\x,5.06)
      {{\footnotesize \txt}};
  }

  \node[panel, minimum width=12.85cm, minimum height=2.18cm] (population) at (7.10,2.91) {};
  \node[title] at (7.10,3.73) {Policy Population over Shared Base Deployments};

  \node[base, minimum width=2.95cm, minimum height=0.82cm] (bases) at (4.29,2.78)
    {{\footnotesize Resident Base}\\[-1pt]
     {\footnotesize Deployments}};

  \node[plainlabel, font=\bfseries\large\sffamily] at (6.23,2.78) {$+$};

  \node[adapter] at (7.29,2.96) {$r_1$};
  \node[adapter] at (8.05,2.96) {$r_2$};
  \node[adapter] at (8.81,2.96) {$r_3$};
  \node[adapter] at (9.57,2.96) {$r_4$};
  \node[adapter] at (10.33,2.96) {$r_5$};
  \node[adapter] at (11.09,2.96) {$\cdots$};

  \node[plainlabel] at (9.19,2.44) {train / evaluate / serve / rollback};

  \node[panel, minimum width=12.85cm, minimum height=2.26cm] (infra) at (7.10,0.34) {};
  \node[title] at (7.10,1.20) {Infrastructure Complexity Hidden Underneath};

  \node[groupbox, minimum width=2.62cm, minimum height=1.10cm, text width=2.44cm] at (2.46,0.15)
    {{\bfseries\scriptsize Scheduler}\\[-1pt]
     {\scriptsize\color{mindlabfg!62}queue / admission}\\[-1pt]
     {\scriptsize\color{mindlabfg!62}placement}};

  \node[groupbox, minimum width=2.62cm, minimum height=1.10cm, text width=2.44cm] at (5.55,0.15)
    {{\bfseries\scriptsize Fault Tolerance}\\[-1pt]
     {\scriptsize\color{mindlabfg!62}records / op state}\\[-1pt]
     {\scriptsize\color{mindlabfg!62}retries}};

  \node[groupbox, minimum width=2.62cm, minimum height=1.10cm, text width=2.44cm] at (8.65,0.15)
    {{\bfseries\scriptsize Adapter Lifecycle}\\[-1pt]
     {\scriptsize\color{mindlabfg!62}restore / export}\\[-1pt]
     {\scriptsize\color{mindlabfg!62}revision visibility}};

  \node[groupbox, minimum width=2.62cm, minimum height=1.10cm, text width=2.44cm] at (11.74,0.15)
    {{\bfseries\scriptsize Serving Residency}\\[-1pt]
     {\scriptsize\color{mindlabfg!62}storage / cache}\\[-1pt]
     {\scriptsize\color{mindlabfg!62}GPU slots}};

  \draw[arrow] (user.south) -- (mint.north);
  \draw[arrow] (mint.south) -- (population.north);
  \draw[arrow] (population.south) -- (infra.north);
\end{tikzpicture}}
    \caption{MinT overview. User intent selects a base model, data and rewards, a LoRA RL recipe, and evaluation or serving targets. MinT turns those inputs into queued work, policy records, and exported revisions, then manages a policy population over shared base deployments while scheduler, fault-tolerance, adapter-lifecycle, and serving-residency mechanisms remain behind the service interface.}
    \label{fig:mint_overview}
\end{figure}

Post-training has become an infrastructure workload. As LLMs move toward trillion-scale deployments and lifelong learning from experience~\citep{yao_second_half_2025,silver_sutton_experience_2025}, frontier teams need reliable frameworks for modern agentic capabilities~\citep{deepseek_v4_release_2026,deepseekv4_2026,glm5_2026,kimi_k25_2026,minimax_m27_2026,qwen35_2026,openai_gpt55_2026,anthropic_opus45_2025}. These workloads combine cluster scheduling, artifact versioning, and frequent deployment. Copying or serving a full checkpoint for each variant no longer scales.

To address these challenges, we propose the MindLab Toolkit (MinT), which uses Low-Rank Adaptation (LoRA) adapters as the basic policy units for post-training and online serving. In MinT, the base model remains resident, while task behaviors, product branches, experimental versions, tenant-specific variants, and rollback points are represented by different adapters~\citep{lora2022,lora_without_regret_2025}. As shown in \Cref{fig:mint_overview}, MinT transforms user intent into queued jobs, policy records, and exported adapter revisions on top of shared base-model deployments. Instead of repeatedly moving, loading, or copying full models between training and serving, the system transfers and manages only compact LoRA parameters. Following the service-interface practice of Tinker~\citep{tinker2025,tinker_cookbook}, MinT connects rollout, update, export, evaluation, serving, and rollback into a LoRA-centered workflow. It hides distributed training, model sharding, tensor-layout conversion, adapter admission, and serving access behind a simple service interface, making large-scale LoRA-based reinforcement learning easier to run, reproduce, and deploy.

\begin{figure}[t]
    \centering
    \resizebox{\textwidth}{!}{\begin{tikzpicture}[x=1cm,y=1cm,>=stealth,font=\footnotesize\sffamily]
  \tikzstyle{title}=[font=\bfseries\small\sffamily, text=mindlabfg, align=center]
  \tikzstyle{row}=[font=\bfseries\footnotesize\sffamily, text=mindlabfg, align=right]
  \tikzstyle{block}=[draw=mindlabfg, thick, rounded corners=0.8mm, fill=white, align=center, inner sep=2pt, minimum width=0.84cm, minimum height=0.54cm]
  \tikzstyle{trainblock}=[block, fill=mindlabblue!22]
  \tikzstyle{adapter}=[draw=mindlabfg, thick, rounded corners=0.7mm, fill=white, align=center, inner sep=1.8pt, minimum width=0.58cm, minimum height=0.34cm]
  \tikzstyle{trainadapter}=[adapter, fill=mindlabblue!22]
  \tikzstyle{note}=[font=\scriptsize\sffamily, text=mindlabfg, align=center]
  \tikzstyle{arrow}=[->, thick, draw=mindlabfg]

  \path[use as bounding box] (-1.45,-4.34) rectangle (14.95,0.42);

  \node[title] at (1.75,0.10) {Full fine-tuning};
  \node[title] at (7.50,0.10) {LoRA merge};
  \node[title] at (13.25,0.10) {MinT multi-LoRA};

  \node[row, anchor=east] at (-0.12,-0.90) {Training\\worker};
  \node[row, anchor=east] at (-0.12,-2.12) {Moved\\file};
  \node[row, anchor=east] at (-0.12,-3.34) {Inference\\worker};

  \foreach \x/\i in {0.65/1,1.75/2,2.85/n} {
    \node[trainblock] (ftw\i) at (\x,-0.90) {$W_{\i}$};
    \node[block] (ftc\i) at (\x,-2.12) {$W_{\i}$};
    \node[block] (fts\i) at (\x,-3.34) {$W_{\i}$};
    \draw[arrow] (ftw\i) -- (ftc\i);
    \draw[arrow] (ftc\i) -- (fts\i);
  }

  \foreach \x/\i in {5.95/1,7.50/2,9.05/n} {
    \node[block] (mb\i) at (\x-0.20,-0.90) {$W$};
    \node[trainadapter] (ml\i) at (\x+0.50,-0.90) {$L_{\i}$};
    \node[block] (mc\i) at (\x,-2.12) {$W'_{\i}$};
    \node[block] (ms\i) at (\x,-3.34) {$W'_{\i}$};
    \coordinate (join\i) at (\x+0.08,-1.18);
    \draw[arrow] (join\i) -- (mc\i);
    \draw[arrow] (mc\i) -- (ms\i);
  }
  \node[block] (mtw) at (12.26,-0.90) {$W$};
  \foreach \x/\i in {12.96/1,13.66/2,14.36/n} {
    \node[trainadapter] (mtl\i) at (\x,-0.90) {$L_{\i}$};
    \node[adapter] (mal\i) at (\x,-2.12) {$L_{\i}$};
    \draw[arrow] (mtl\i) -- (mal\i);
  }

  \node[block] (miw) at (12.26,-3.34) {$W$};
  \foreach \x/\i in {12.96/1,13.66/2,14.36/n} {
    \node[adapter] (msl\i) at (\x,-3.34) {$L_{\i}$};
    \draw[arrow] (mal\i) -- (msl\i);
  }

  \node[note] (legend) at (6.75,-4.07) {\quad blue = trained parameters; $W$ = base model; $L_i$ = LoRA adapter; $W'_i$ = merged checkpoint};
  \node[trainadapter, minimum width=0.46cm, minimum height=0.26cm, anchor=east] at (legend.west) {};
\end{tikzpicture}}
    \caption{What crosses from training to serving under three adaptation paths. Blue marks the trained parameters. Full fine-tuning moves a full checkpoint for each variant. Merge-based LoRA trains adapters beside a resident base, then moves merged full checkpoints. MinT trains adapters beside a resident base and moves only adapter revisions to an inference engine that already holds the compatible base.}
    \label{fig:mint_handoff_paths}
\end{figure}
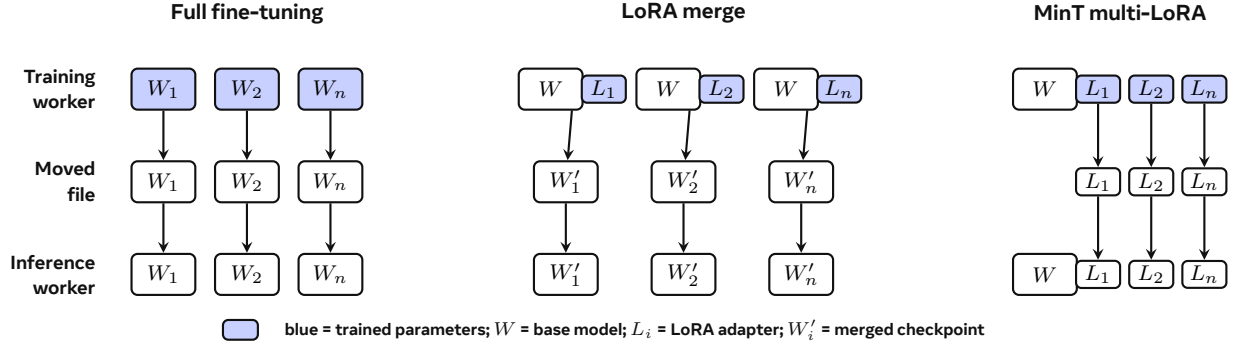

This adapter-centered design changes what crosses the training-serving boundary. Full fine-tuning moves a full checkpoint for each trained variant. Merge-based LoRA reduces training memory, but still folds the adapter back into the base model and moves a merged checkpoint before inference. MinT instead exports the updated LoRA as a serving-compatible adapter revision, checks that it matches the resident base model, and loads it into an inference engine that already holds that base. \Cref{fig:mint_handoff_paths} illustrates this difference: MinT moves adapter revisions, not full model checkpoints.

In this report, we introduce MinT's capabilities along three scaling axes. \textit{Scale Up} supports LoRA RL on frontier-scale dense and MoE architectures, with model-parallel training and serving paths validated beyond 1T total parameters~\citep{qwen3_2025,deepseekv4_2026,glm5_2026,kimi_k2_2025}. This axis keeps the same adapter-revision path usable when the base requires tensor-parallel or expert-parallel placement, sparse-route consistency, and distributed adapter export. \textit{Scale Down} minimizes the training-serving handoff by moving only the exported LoRA adapter, which can be less than 1\% of the base-model size in compact rank-1 settings, eliminating full-checkpoint materialization entirely. In our measurements, adapter-only handoff reduces the measured handoff step by $18.3\times$ on a 4B dense model and by $2.85\times$ on a 30B MoE model; under the same resident-base allocation, concurrent multi-policy GRPO shortens wall time by $1.77\times$ and $1.45\times$, respectively, without increasing peak memory. \textit{Scale Out} expands the policy namespace while keeping engine-local execution bounded. MinT separates durable policy addressability from CPU/GPU hot working sets, builds on modern multi-adapter serving support~\citep{vllm2023,punica2024,slora2023,dlora2024,loraserve2025}, and treats first-touch adapter loading as scheduled service work. In the updated 1M packed-catalog experiment, MinT materializes 1{,}000{,}000 adapter revisions with zero build errors and validates a 256-adapter audit across all 100 shards; the serving measurements then operate selected revisions from that catalog through bounded CPU-cache and GPU-batch working sets. This changes the scale-out claim from a capacity sketch to a measured artifact: the million-entry catalog is real, while local residency remains bounded. The follow-up probes also identify the system work that must be optimized. Catalog registration is fast, but cold activation can interfere with old warm tenants, and fragmented MoE LoRA tensors make the load path object-bound rather than byte-bound. Packed MoE LoRA tensors improve live engine loading by $8.5$--$8.7\times$. Admission-aware readiness gates expose a new adapter only after background prewarm, so ready-path user requests no longer perform adapter loading; the tradeoff is rollout delay before the adapter becomes selectable. Together, these axes make continuously evolving LoRA post-training cheaper to operate, easier to reproduce, and better matched to multi-tenant policy populations.

MinT contributes four concrete pieces:

\begin{itemize}
    \item \textbf{Adapter lifecycle from update to serving.} MinT exports trained LoRAs as PEFT adapter revisions, including sharded-to-serving conversion, compatibility checks, rollout records, sampler loading, served-result attribution, and rollback. The adapter revision becomes the fixed behavior selected by rollout, evaluation, online serving, and recovery.

    \item \textbf{Large-scale multi-LoRA RL training.} MinT time-slices LoRA training sessions over resident shared bases and supports both single-worker PEFT and distributed Megatron training paths. It reports end-to-end LoRA RL on large dense and MoE deployments, including 235B-A22B-scale training and a 1T-class MoE countdown-task path.

    \item \textbf{Policy-population multi-LoRA serving.} MinT serves exported adapters through shared-base vLLM engines, separates durable policy addressability from CPU/GPU hot working sets, and treats cold loading as scheduled work for cache misses. Its 1M packed-catalog experiment materializes and audits a million adapter revisions, while packed MoE LoRA representation and admission-aware readiness gates convert two measured bottlenecks -- tensor fanout and cold-activation interference -- into explicit serving controls.

    \item \textbf{Public reproducibility paths.} MinT provides a Tinker-compatible API~\citep{tinker2025,tinker_cookbook} and uses mint-cookbook recipes~\citep{mint_cookbook2026} to reproduce SFT, preference optimization, rollout-based RL, and AutoResearch examples through the same adapter lifecycle.
\end{itemize}

\section{From LoRA Adapters to Managed Policy Revisions}

Once many trained behaviors share a few resident base models, MinT has to separate the bytes that execute a behavior from the service state that makes the behavior manageable. An \emph{adapter revision} is a fixed, exported snapshot of the LoRA adapter, frozen at a specific training step and stored in serving tensor layout; it is the executable LoRA payload that crosses from training to rollout, evaluation, and serving (it is not a training iteration or a transfer event). The policy record is the service-owned lifecycle state that makes that payload reproducible, reloadable, and rollbackable. This distinction lets a shared base support many LoRA policies without turning every policy into another full checkpoint or another full-model server.

An RL update leaves more than a LoRA file. After one update, the trainer has adapter tensors, optimizer moments, scheduler position, accumulated gradients, and rollout records that may still be needed for scoring. A sampler cannot consume that training checkpoint directly. It needs a fixed adapter revision in serving tensor layout, tied to the base model that is already resident in the inference engine. Later requests may select the same revision after the original worker has exited or after the serving cache has evicted the adapter bytes.

\begin{wrapfigure}{r}{0.54\textwidth}
    \vspace{-0.7\baselineskip}
    \centering
    \resizebox{\linewidth}{!}{\begin{tikzpicture}[x=1cm,y=1cm,>=stealth,font=\small\sffamily]
  \definecolor{mintauxteal}{HTML}{009E9A}
  \definecolor{mintauxviolet}{HTML}{7057D2}
  \definecolor{mintauxamber}{HTML}{D98B00}
  \colorlet{loraone}{mindlabblue!85}
  \colorlet{loratwo}{mintauxteal!82}
  \colorlet{lorathree}{mintauxviolet!76}
  \colorlet{lorafour}{mindlabblue!58}
  \colorlet{lorafive}{mintauxamber!78}
  \colorlet{lorasix}{mindlabblue!38}
  \colorlet{loraseven}{mintauxteal!48}
  \colorlet{loraeight}{mindlabfg!30}
  \colorlet{coldtier}{mindlabbluepale!35}
  \colorlet{warmtier}{mintauxteal!16}
  \colorlet{hottier}{mindlabbluepale!78}
  \colorlet{coolone}{mindlabblue!18}
  \colorlet{cooltwo}{mintauxteal!24}
  \colorlet{coolthree}{mintauxviolet!24}
  \colorlet{coolfour}{mindlabblue!30}
  \colorlet{coolfive}{mintauxamber!28}
  \colorlet{warmone}{mindlabblue!38}
  \colorlet{warmtwo}{mintauxteal!42}
  \colorlet{warmthree}{mintauxviolet!36}
  \colorlet{warmfour}{mindlabblue!28}
  \colorlet{hotone}{mindlabblue!85}
  \colorlet{hottwo}{mintauxteal!72}
  \colorlet{hotthree}{mintauxviolet!64}
  \tikzstyle{frame}=[draw=mindlabfg, very thick, rounded corners=1mm, fill=white]
  \tikzstyle{tier}=[draw=mindlabfg, thick, rounded corners=0.8mm, fill=white]
  \tikzstyle{base}=[draw=mindlabfg, thick, rounded corners=1mm, fill=mindlabbluepale!45, align=center, inner sep=2pt, minimum width=0.92cm, minimum height=0.30cm]
  \tikzstyle{adapter}=[draw=mindlabfg, thick, rounded corners=1mm, minimum width=0.78cm, minimum height=0.22cm, inner sep=0pt]
  \tikzstyle{arrow}=[->, very thick, draw=mindlabfg, shorten >=2pt, shorten <=2pt]
  \tikzstyle{returnarrow}=[->, thick, dashed, draw=mindlabfg, shorten >=2pt, shorten <=2pt]
  \tikzstyle{title}=[font=\bfseries\small\sffamily, text=mindlabfg, align=center]
  \tikzstyle{label}=[font=\scriptsize\sffamily, text=mindlabfg, align=center, fill=white, inner sep=1.2pt]
  \tikzstyle{smalllabel}=[font=\scriptsize\sffamily, text=mindlabfg, align=center, fill=white, inner sep=1pt]
  \tikzstyle{plainlabel}=[font=\scriptsize\sffamily, text=mindlabfg, align=center, inner sep=1pt]
  \tikzstyle{tierlabel}=[font=\scriptsize\sffamily, text=mindlabfg, align=center, inner sep=0pt]

  \path[use as bounding box] (1.05,-3.18) rectangle (8.95,2.18);

  \node[frame, minimum width=2.42cm, minimum height=1.08cm] (trainer) at (2.95,1.00) {};
  \node[title] at (2.95,1.77) {Training worker};
  \node[base] at (2.50,1.00) {Base};
  \node[adapter, fill=hotone] at (3.48,1.02) {};

  \node[frame, minimum width=2.42cm, minimum height=1.08cm] (rollout) at (7.60,1.00) {};
  \node[title] at (7.60,1.77) {Rollout worker};
  \node[base] at (7.15,1.00) {Base};
  \node[adapter, fill=hotone] at (8.13,1.30) {};
  \node[adapter, fill=hottwo] at (8.13,1.00) {};
  \node[adapter, fill=hotthree] at (8.13,0.70) {};

  \node[frame, fill=coldtier, minimum width=7.07cm, minimum height=2.12cm] (population) at (5.275,-1.55) {};
  \node[tierlabel, anchor=west] at (1.97,-0.78) {cool: stored};
  \foreach \x/\y/\c in {
	    2.36/-1.06/coolone,2.36/-1.36/cooltwo,2.36/-1.66/coolthree,2.36/-1.96/coolfour,2.36/-2.26/coolfive,
	    3.21/-1.06/cooltwo!82,3.21/-1.36/coolfour!86,3.21/-1.66/coolfive!82,3.21/-1.96/coolone!74,3.21/-2.26/coolthree!76
  } {
    \node[adapter, fill=\c, minimum width=0.78cm] at (\x,\y) {};
  }
  \node[plainlabel] at (3.21,-2.50) {$\cdots$};

  \node[tier, fill=warmtier, minimum width=4.555cm, minimum height=1.60cm] (host) at (6.2775,-1.70) {};
  \node[tierlabel, anchor=west] at (4.18,-1.13) {warm: CPU cache};
  \foreach \x/\y/\c in {
	    4.57/-1.40/warmone,4.57/-1.70/warmtwo,4.57/-2.00/warmthree,
	    5.42/-1.40/warmfour,5.42/-1.70/warmone!82,5.42/-2.00/warmtwo!82
  } {
    \node[adapter, fill=\c, minimum width=0.78cm] at (\x,\y) {};
  }

  \node[tier, fill=hottier, minimum width=1.88cm, minimum height=1.18cm] (gpu) at (7.36,-1.79) {};
  \node[tierlabel, anchor=west] at (6.55,-1.35) {hot: GPU batch};
  \node[adapter, fill=hotone] at (6.94,-1.59) {};
  \node[adapter, fill=hottwo] at (6.94,-1.89) {};
  \node[adapter, fill=hotthree] at (6.94,-2.19) {};
  \node[title] at (5.275,-2.88) {Policy population};

  \draw[arrow] (trainer.south) -- node[label, right, pos=0.50, xshift=2pt] {save revision} (trainer.south |- population.north);
  \draw[arrow] (population.north -| rollout.south) -- node[label, left, pos=0.50, xshift=-2pt] {selected adapter} (rollout.south);
  \draw[returnarrow] (rollout.west) -- node[label, above, yshift=2pt] {rollout records} (trainer.east);
\end{tikzpicture}}
    \caption{Adapter lifecycle in MinT. Training saves revisions into the policy population; rollout uses the hot GPU-batch subset and returns records for the next update.}
    \label{fig:policy_lifecycle}
    \vspace{-0.6\baselineskip}
\end{wrapfigure}
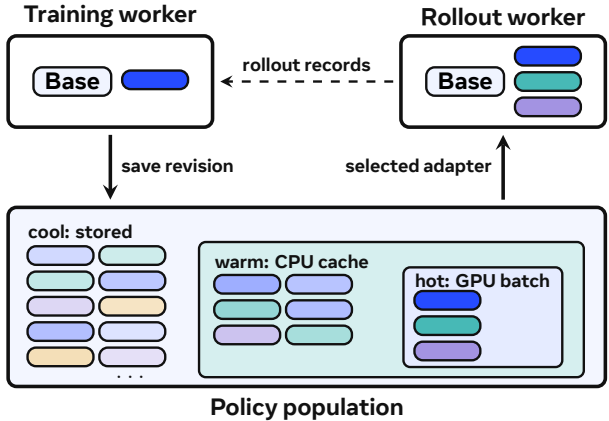

The service therefore has to answer four concrete questions after every update. Which base deployment can run this adapter? Which training checkpoint should a trainer restore if training resumes? Which fixed adapter revision should rollout, evaluation, or serving use? Where are those adapter bytes now: active in a GPU batch, cached in CPU memory, or only present in shared storage? MinT stores these answers separately because they change at different times.

A MinT policy record is the durable entry for one trained behavior over a compatible base model. It names the base version, LoRA rank and target modules, the latest training checkpoint, the rollout records kept for later updates, and the exported adapter revisions available for fixed behavior. A live training session is a temporary restoration of that record on a worker. MinT allows one training session per policy record; concurrent branches become separate records so each worker writes a different adapter and optimizer state. Export freezes the current training checkpoint into a fixed adapter revision. Evaluation, rollout, and online serving select a revision, while worker placement and cache state can change around it.

The RL loop crosses this lifecycle on every iteration. Rollout samples trajectories with a selected adapter revision on an inference engine that already holds the base. Training recomputes token probabilities for those trajectories, applies the objective, and changes the adapter and optimizer state. The next rollout needs another export because inference engines consume fixed PEFT adapter files in serving layout. Trainer-local optimizer checkpoints stay on the training side. Evaluation uses the same fixed-revision rule so a reported score names the adapter revision that produced it.

Serving adds another time scale. A request can name an adapter revision long after its bytes have left the local engine. The request first resolves a user-facing policy name to an exported revision and an engine-local adapter id. If the adapter is already active in the GPU batch, decoding can use it immediately. If it is cached in CPU memory, the engine promotes it. If it is absent from the local cache, the serving actor fetches the adapter file from shared storage and loads it before decoding. This lets MinT keep a large addressable catalog while bounding the much smaller CPU-cache and same-batch working sets.

Training has the opposite problem: one resident base deployment may serve several policy records over time. When a trainer switches from policy $A$ to policy $B$, MinT must save $A$'s adapter, optimizer moments, scheduler position, accumulated gradients, and unconsumed rollout records before restoring $B$'s corresponding state. The base weights stay resident. Only the LoRA tensors and training state change. Policies may also use different LoRA ranks or target-module sets, so the policy record carries the adapter shape that the worker must allocate and restore.

Sparse models split rollout metadata into two cases. MoE rollout records can carry selected expert ids. When the training backend can map those ids to its expert-parallel layout, log-probability scoring reuses the recorded expert path; when the ids are missing or unmappable, MinT removes that token from the replayed policy-gradient term. GLM-5-style dynamic sparse attention uses a different treatment. MinT currently lacks replay for every DSA indexer selection. The rollout record stores the backend/model path and correction policy, not per-token DSA indexer selections, and IcePop-style rollout correction zeroes importance weights whose training/rollout ratio falls outside the trusted band. That mitigation keeps unstable tokens out of the gradient term; it does not reconstruct the exact sparse-attention token set selected by the inference engine.

The adapter revision is the behavior-carrying payload; the policy record is the service state that makes that payload schedulable and durable. In this paper, a deployable LoRA policy over a resident base is the LLM behavior that MinT trains, evaluates, serves, and rolls back. MinT keeps the base deployment, training checkpoint, rollout record, exported adapter revision, and adapter cache state as separate facts so one trained behavior can be resumed, scored, exported, evicted, reloaded, and served in a larger population.

\section{System Design}
\label{sec:architecture}

MinT is a Tinker-compatible managed service for LoRA RL over resident base-model deployments. A client package calls the service to sample rollouts, compute gradients, apply an optimizer step, export an adapter revision, evaluate or serve that revision, and poll for the result. The service keeps the user-facing loop simple while dense and MoE bases remain loaded in resident trainers, samplers, and serving actors. Durable operation ids and policy records keep retries precise: the service knows which adapter generated a rollout, which checkpoint can resume training, which export is visible to samplers, and which resident worker can run the next request.

\Cref{fig:mint_overview} gives the service view, and \cref{fig:mint_handoff_paths} shows the training-serving boundary. The runtime must preserve both: a simple client interface and an adapter-only handoff. MinT therefore separates service control from resident compute. The service plane validates each call, records a pollable operation id, resolves the policy record or exported revision, and admits work onto a compatible worker. The compute plane has three service roles because different operations on one policy record require different compatible worker shapes. Single-worker PEFT trainers run LoRA updates when one worker can hold the base replica. Distributed Megatron trainer groups run LoRA updates when tensor, pipeline, or expert parallelism partitions the base and adapter tensors across ranks. vLLM sampler and serving actors hold inference bases and attach exported LoRA adapters for rollout, evaluation, and online serving. The runtime path has four mechanisms: service-plane visibility and policy-record resolution, time-sliced training over resident bases, export from trainer state to serving adapter files, and shared-base rollout or serving with adapter cache tiers.

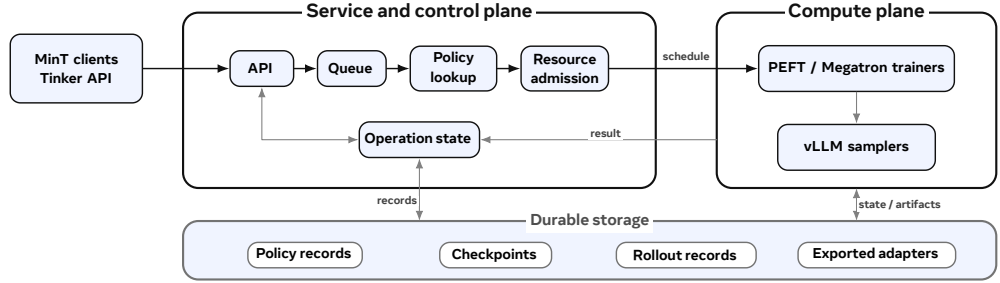
\begin{figure}[tbp]
    \centering
    \resizebox{\textwidth}{!}{\usetikzlibrary{arrows.meta,calc,fit,positioning}
\colorlet{mintdark}{mindlabfg}
\colorlet{mintfill}{mindlabbluepale!45}
\begin{tikzpicture}[
    font=\sffamily\small,
    >=Latex,
    x=1cm,
    y=1cm,
    line cap=round,
    line join=round,
    panel/.style={
        draw=mintdark,
        very thick,
        rounded corners=3mm,
        fill=white
    },
    box/.style={
        draw=mintdark,
        thick,
        rounded corners=2mm,
        fill=mintfill,
        align=center,
        inner sep=3pt,
        minimum height=0.62cm
    },
    storagebox/.style={
        draw=black!55,
        thick,
        rounded corners=2mm,
        fill=white,
        align=center,
        inner sep=3pt,
        minimum height=0.50cm
    },
    patharrow/.style={
        ->,
        thick,
        draw=mintdark
    },
    softarrow/.style={
        ->,
        semithick,
        draw=black!50
    },
    botharrow/.style={
        <->,
        semithick,
        draw=black!50
    },
    statepath/.style={
        -,
        semithick,
        draw=black!45
    },
    metapath/.style={
        -,
        semithick,
        dashed,
        draw=black!50
    },
    label/.style={
        font=\sffamily\scriptsize,
        fill=white,
        inner sep=1pt,
        text=black!75,
        align=center
    },
    sublabel/.style={
        font=\sffamily\tiny,
        text=black!62,
        align=center
    },
    paneltitle/.style={
        font=\sffamily\bfseries\large,
        text=mintdark,
        fill=white,
        inner sep=2pt
    },
    storagetitle/.style={
        font=\sffamily\bfseries\normalsize,
        text=black!70,
        fill=white,
        inner sep=2pt
    },
    note/.style={
        font=\sffamily\scriptsize,
        text=black!65,
        align=center
    }
]

\path[use as bounding box] (-2.00,0.15) rectangle (20.80,5.92);


\node[
    box,
    minimum width=2.60cm,
    minimum height=1.35cm,
    text width=2.30cm
] (clients) at (-0.70,4.55) {
    MinT clients\\
    Tinker API
};

\draw[panel] (1.40,2.20) rectangle (10.70,5.65);
\node[paneltitle] at (6.05,5.65) {Service and control plane};

\draw[panel] (11.90,2.20) rectangle (17.40,5.65);
\node[paneltitle] at (14.65,5.65) {Compute plane};

\draw[
    draw=black!55,
    very thick,
    rounded corners=3mm,
    fill=mintfill
] (1.40,0.40) rectangle (17.40,1.55);

\node[storagetitle] at (9.40,1.55) {Durable storage};


\node[
    box,
    minimum width=1.25cm,
    minimum height=0.70cm
] (api) at (2.95,4.55) {
    API
};

\node[
    box,
    minimum width=1.35cm,
    minimum height=0.70cm
] (queue) at (4.72,4.55) {
    Queue
};

\node[
    box,
    minimum width=1.70cm,
    minimum height=0.88cm,
    text width=1.45cm
] (lookup) at (6.72,4.55) {
    Policy\\
    lookup
};

\node[
    box,
    minimum width=1.72cm,
    minimum height=0.88cm,
    text width=1.50cm
] (admit) at (8.91,4.55) {
    Resource\\
    admission
};

\node[
    box,
    minimum width=2.15cm,
    minimum height=0.70cm
] (future) at (6.05,3.15) {
    Operation state
};


\node[
    box,
    minimum width=3.85cm,
    minimum height=0.90cm,
    text width=3.60cm
] (trainers) at (14.65,4.55) {
    PEFT / Megatron trainers
};

\node[
    box,
    minimum width=3.10cm,
    minimum height=0.90cm,
    text width=2.85cm
] (samplers) at (14.65,3.00) {
    vLLM samplers
};


\node[
    storagebox,
    minimum width=2.25cm
] (policy_store) at (3.80,0.88) {
    Policy records
};

\node[
    storagebox,
    minimum width=2.20cm
] (ckpt) at (7.55,0.88) {
    Checkpoints
};

\node[
    storagebox,
    minimum width=2.75cm
] (rollout) at (11.30,0.88) {
    Rollout records
};

\node[
    storagebox,
    minimum width=3.10cm
] (adapter) at (15.05,0.88) {
    Exported adapters
};


\draw[patharrow]
    (clients.east) -- (api.west);

\draw[patharrow] (api.east) -- (queue.west);
\draw[patharrow] (queue.east) -- (lookup.west);
\draw[patharrow] (lookup.east) -- (admit.west);

\draw[patharrow]
    (admit.east) -- (trainers.west);
\node[label] at (11.30,4.79) {schedule};


\draw[botharrow]
    (api.south) |- (future.west);

\draw[softarrow]
    (11.90,3.15) -- node[label, above, pos=0.47] {result} (future.east);


\draw[softarrow]
    (trainers.south) -- (samplers.north);

\draw[botharrow]
    (14.65,2.20) -- node[label, right, pos=0.50] {state / artifacts} (14.65,1.55);

\draw[botharrow]
    (future.south) -- node[label, left, pos=0.68] {records} (6.05,1.55);

\end{tikzpicture}}
\caption{MinT runtime organization. Tinker-compatible client packages call the service and receive a pollable operation id. The service queues the operation, reads the policy record, admits work onto the compute plane, and records completed operation results until clients poll them. Trainer workers update LoRA tensors and optimizer state over a resident base model; vLLM sampler and serving actors generate with exported adapter revisions. Durable storage holds checkpoints, rollout records, and exported revisions.}
    \label{fig:mint_architecture}
\end{figure}

\subsection{Service Plane and Resident Workers}

The service plane owns the client-visible state for each operation. It decides when a request becomes scheduled work, when a produced file becomes a usable checkpoint, exported revision, or result, and what a retry observes after a worker exits.

\paragraph{Operation visibility.}
The service plane turns each client call into scheduled work on a compatible resident worker. It validates the request, enqueues the operation, returns an id that the client can poll, and admits the work only when a worker with the required base deployment and adapter capacity is available. A worker may keep temporary tensors while it executes the call. A checkpoint, exported adapter revision, rollout record, or operation result becomes visible only when MinT writes the metadata entry that names its completed files and stores the pollable result. If a worker crashes after writing adapter files but before recording that entry, later requests cannot select those uncommitted files; the caller can retry the operation, and cleanup can remove unreferenced attempt paths. The same visibility rule covers training checkpoints, exported adapter revisions, rollout records, and pollable operation results.

\paragraph{Policy record resolution.}
Every training or sampling call resolves to a policy record before it reaches a worker. The record names the compatible base version, LoRA rank, target modules, and storage locations for checkpoints, rollout records, optimizer state, and exported revisions. Training restores the state named by the record: adapter tensors, optimizer moments, scheduler position, gradients, and rollout metadata. Sampling selects an exported adapter revision from the same record. If the service process restarts, completed checkpoints, exported revisions, rollout records, and finished operation results remain recoverable from storage.

\paragraph{Worker admission and eviction.}
Training and sampling workers consume cluster GPUs in different shapes, so the service admits them through one resource view. A single-worker PEFT trainer occupies one model replica. A Megatron training group spans tensor-parallel, pipeline-parallel, or expert-parallel ranks. A vLLM sampler reserves memory for a base model plus adapter slots. MinT tracks live workers, active training sessions, in-flight generation, pinned adapters, idle time, and reclaimable base deployments. Evicting an idle trainer frees compute while stored LoRA tensors, optimizer state, rollout records, and exported revisions remain requestable. Evicting an idle sampler removes actor-local cached adapters and GPU-batch slots while the exported revisions remain in shared storage for later loading.

\begin{figure}[t]
    \centering
    \resizebox{0.78\textwidth}{!}{\begin{tikzpicture}[x=1cm,y=1cm,>=stealth,font=\scriptsize\sffamily]
  \tikzstyle{actor}=[draw=mindlabfg, thick, rounded corners=1mm, fill=white, align=center]
  \tikzstyle{base}=[draw=mindlabfg, thick, rounded corners=0.8mm, fill=mindlabbluepale!45, align=center, inner sep=3pt]
  \tikzstyle{slot}=[draw=mindlabfg, thick, rounded corners=0.8mm, fill=mindlabblue!18, align=center, inner sep=3pt]
  \tikzstyle{switch}=[draw=mindlabfg, thick, rounded corners=0.8mm, fill=white, align=center, inner sep=3pt]
  \tikzstyle{store}=[draw=mindlabfg, thick, rounded corners=0.8mm, fill=white, align=center, inner sep=3pt]
  \tikzstyle{arrow}=[->, thick, draw=mindlabfg, shorten >=1pt, shorten <=1pt]
  \tikzstyle{both}=[<->, thick, draw=mindlabfg, shorten >=1pt, shorten <=1pt]
  \tikzstyle{label}=[font=\scriptsize\sffamily, align=center, fill=white, inner sep=1pt]
  \tikzstyle{title}=[font=\bfseries\footnotesize\sffamily, align=center, text=mindlabfg]

  \path[use as bounding box] (-0.20,-2.35) rectangle (10.15,1.85);

  \draw[arrow] (0.70,1.34) -- (9.45,1.34);
  \node[label] at (1.30,1.58) {$t_1$};
  \node[label] at (4.92,1.58) {$t_2$};
  \node[label] at (8.55,1.58) {$t_3$};

  \node[actor, minimum width=8.90cm, minimum height=1.70cm] (trainer) at (4.92,0.35) {};
  \node[title] at (4.92,1.05) {one trainer actor};
  \node[base, minimum width=7.95cm, minimum height=0.36cm] (base) at (4.92,-0.16) {resident base model};

  \node[slot, minimum width=1.05cm, minimum height=0.38cm] (a1) at (1.30,0.45) {LoRA A};
  \node[switch, minimum width=1.08cm, minimum height=0.46cm] (sw1) at (3.10,0.45) {switch};
  \node[slot, minimum width=1.05cm, minimum height=0.38cm] (b2) at (4.92,0.45) {LoRA B};
  \node[switch, minimum width=1.08cm, minimum height=0.46cm] (sw2) at (6.75,0.45) {switch};
  \node[slot, minimum width=1.05cm, minimum height=0.38cm] (a3) at (8.55,0.45) {LoRA A};

  \node[store, minimum width=4.35cm, minimum height=0.62cm] (records) at (4.92,-1.48)
    {policy records\\\scriptsize LoRA tensors, optimizer, gradients, rollouts};

  \draw[arrow] (a1.east) -- (sw1.west);
  \draw[arrow] (sw1.east) -- (b2.west);
  \draw[arrow] (b2.east) -- (sw2.west);
  \draw[arrow] (sw2.east) -- (a3.west);

  \draw[both] (sw1.south) -- (records.north -| sw1.south);
  \node[label, anchor=east] at (3.02,-0.78) {save A / load B};
  \draw[both] (sw2.south) -- (records.north -| sw2.south);
  \node[label, anchor=west] at (6.82,-0.78) {save B / load A};
\end{tikzpicture}}
    \caption{Time-sliced multi-LoRA training. One trainer keeps the base model resident. A scheduled policy loads its LoRA tensors, optimizer state, gradients, and rollout records into the trainer; the previous policy writes the same state back to storage before the switch.}
    \label{fig:mint_training_state_swap}
\end{figure}
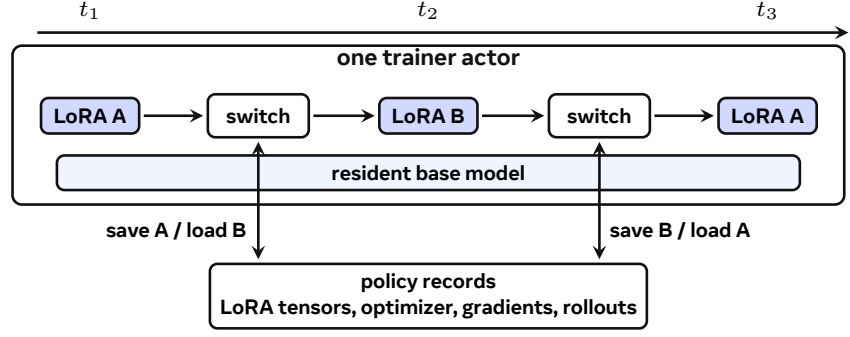

\subsection{Time-Sliced Multi-LoRA Training}

MinT trains many LoRA policies over one resident base without allocating one base replica per policy. Replicating the frozen base for every active policy would spend most GPU memory on repeated weights and would remove the memory advantage that made LoRA attractive. Multi-LoRA training kernels, as explored by mLoRA for fine-tuning workloads~\citep{mlora2024}, can update several adapters in one batch. MinT instead time-slices policies on each trainer because RL requests usually contain enough rollout tokens to form efficient per-policy batches. The trainer swaps only the selected policy's LoRA tensors and optimizer state between requests while the frozen base remains loaded.

Each trainer time-slices locally. The service can run several resident trainers and samplers for the same base family, and the queue assigns ready policy operations across that pool. Each trainer executes one policy training session at a time so optimizer state and accumulated gradients have one writer, while other policies can progress on other workers or in sampler actors.

When policy $A$ yields the trainer to policy $B$, MinT writes $A$'s training state out and restores $B$'s state before the next gradient computation. The swapped state includes LoRA tensors, optimizer moments, scheduler position, accumulated gradients, and rollout records. After the update, MinT writes the changed state back. The base model stays in GPU memory across policy switches, while inactive adapter tensors and optimizer state can sit in CPU memory or storage until another request selects them.

The service supports different LoRA ranks and target-module sets on the same resident base when the worker was configured for those shapes. Two policies may use different ranks or target different modules, such as attention-only adapters versus adapters on attention, MLP, and output layers. A worker advertises supported adapter-shape limits when it joins the service. Within those limits, the trainer allocates adapter slots at configured maximum shapes, pads smaller adapters into those slots, and masks inactive rows or modules so each policy updates only its own parameters. Requests outside those limits require a worker configured for the larger rank or target-module set.

Single-worker PEFT trainers and distributed Megatron trainer groups restore LoRA tensors, run the update, and checkpoint the result. For smaller dense models, a trainer restores one adapter into one complete model replica. For Megatron MoE models, a trainer group restores tensor-parallel slices and expert-parallel adapter tensors onto the ranks that own the corresponding base shards. The model-parallel base remains distributed and resident; only LoRA tensors and optimizer state change when the scheduler selects a different policy.

\subsection{Adapter Data Flow Between Training and Serving}

After the trainer updates a LoRA, the next rollout, evaluation run, or serving request must use that adapter on an inference engine. The trainer holds LoRA tensors, optimizer state, and, for distributed training, rank-local tensor shards. vLLM expects a fixed adapter revision in the serving tensor layout, with optimizer state and rank-local training files removed. MinT exports the trained LoRA in PEFT format, carrying adapter tensors, rank, target modules, and base-model compatibility metadata without copying base checkpoint bytes.

Distributed export rebuilds the serving adapter from sharded trainer files. Tensor-parallel ranks may hold slices of the same adapter tensor, and expert-parallel ranks may own different expert adapters. MinT gathers tensor-parallel slices, writes replicated tensors once, collects expert tensors from the ranks that own experts, and emits the PEFT layout expected by the sampler. For MoE adapters with shared-expert LoRA on each expert-parallel shard, the export path deduplicates shared-expert tensors, and the loader materializes one copy instead of repeating it for every shard. The file that crosses from training to rollout or serving is one adapter revision assembled in serving layout; it excludes merged base weights and rank-local training files.

The sampler admits an exported adapter only when its base model family, target modules, rank, and tensor layout match the resident base deployment and configured adapter buffers. The policy may continue training after export, while evaluation and serving select the fixed revision produced by a particular export.

\begin{figure}[H]
    \centering
    \resizebox{0.84\textwidth}{!}{\begin{tikzpicture}[x=1cm,y=1cm,>=stealth,font=\footnotesize\sffamily]
  \tikzstyle{actor}=[draw=mindlabfg, very thick, rounded corners=1mm, fill=white]
  \tikzstyle{box}=[draw=mindlabfg, thick, rounded corners=0.8mm, fill=mindlabbluepale!45, align=center, inner sep=4pt]
  \tikzstyle{store}=[draw=mindlabfg, thick, rounded corners=0.8mm, fill=white, align=center, inner sep=4pt]
  \tikzstyle{slotbox}=[draw=mindlabfg, thick, rounded corners=0.8mm, fill=white, align=center, inner sep=3pt]
  \tikzstyle{adapter}=[draw=mindlabfg, thick, rounded corners=1mm, align=center, minimum width=0.86cm, minimum height=0.23cm, inner sep=0pt, fill=mindlabfg!8]
  \tikzstyle{selected}=[draw=mindlabfg, thick, rounded corners=1mm, align=center, font=\scriptsize\sffamily, minimum width=1.14cm, minimum height=0.30cm, inner sep=1.2pt, fill=mindlabblue!38]
  \tikzstyle{arrow}=[->, very thick, draw=mindlabfg, shorten >=2pt, shorten <=2pt]
  \tikzstyle{hot}=[->, very thick, draw=hotpath, shorten >=2pt, shorten <=2pt]
  \tikzstyle{cold}=[->, thick, dashed, draw=coldpath, shorten >=2pt, shorten <=2pt]
  \tikzstyle{label}=[font=\footnotesize\sffamily, align=center, fill=white, inner sep=1.5pt, text=mindlabfg]
  \tikzstyle{smalllabel}=[font=\scriptsize\sffamily, align=center, fill=white, inner sep=1.1pt, text=mindlabfg]
  \tikzstyle{title}=[font=\bfseries\footnotesize\sffamily, align=center, text=mindlabfg]
  \colorlet{hotpath}{mindlabblue}
  \colorlet{coldpath}{mindlabline}

  \path[use as bounding box] (0.20,-2.42) rectangle (14.05,2.18);

  \node[box, minimum width=1.92cm, minimum height=0.72cm] (map) at (2.92,0.40) {Serving map\\policy $\to r$};

  \node[actor, minimum width=7.02cm, minimum height=3.70cm] (actor) at (8.53,-0.18) {};
  \node[title] at (8.53,1.44) {vLLM sampler actor};

  \node[slotbox, minimum width=1.86cm, minimum height=1.28cm] (gpu) at (6.64,0.40) {};
  \node[smalllabel, anchor=south] at (6.64,1.09) {GPU LoRA slots};
  \node[adapter] at (6.64,0.72) {};
  \node[selected] (selected) at (6.64,0.40) {$L_r$};
  \node[adapter] at (6.64,0.08) {};

  \node[box, minimum width=2.10cm, minimum height=0.76cm] (infer) at (10.30,0.40)
    {vLLM\\generation};
  \node[store, minimum width=2.10cm, minimum height=0.78cm] (base) at (10.30,-1.35)
    {resident\\model};
  \node[label, anchor=west] (tokens) at (12.54,0.40) {response tokens};

  \node[store, minimum width=1.86cm, minimum height=0.78cm] (warm) at (6.64,-1.35)
    {CPU adapter\\cache};
  \node[store, minimum width=2.50cm, minimum height=0.78cm] (catalog) at (2.63,-1.35)
    {shared adapter\\storage};

  \draw[arrow] (0.62,0.40) -- node[smalllabel, above, pos=0.42] {request} (map.west);
  \draw[hot] (map.east) -- node[smalllabel, above, yshift=3pt, pos=0.50, text=hotpath] {hit: $L_r$ resident} (gpu.west);
  \draw[hot] (gpu.east) -- node[smalllabel, above, yshift=3pt, pos=0.50, text=hotpath] {selected $L_r$} (infer.west);
  \draw[arrow] (base.north) -- (infer.south);
  \draw[arrow] (infer.east) -- (tokens.west);

  \draw[cold] (catalog.east) -- node[smalllabel, above, pos=0.50, text=coldpath] {fetch $r$} (warm.west);
  \draw[cold] (warm.north) -- node[smalllabel, right, xshift=6pt, pos=0.44, text=coldpath] {admit $L_r$} (gpu.south);

  \draw[hot, -] (12.42,-1.09) -- (12.88,-1.09);
  \node[smalllabel, anchor=west, text=hotpath] at (12.94,-1.09) {hot path};
  \draw[cold, -, dashed] (12.42,-1.61) -- (12.88,-1.61);
  \node[smalllabel, anchor=west, text=coldpath] at (12.94,-1.61) {cold path};
\end{tikzpicture}}
\caption{Shared-base multi-LoRA sampling. A request reaches the serving map, which resolves the selected policy to exported adapter revision $r$. The solid path uses revision $r$ because its LoRA $L_r$ is already in a GPU batch slot. The vLLM actor runs inference with the resident base and $L_r$, then returns response tokens. Dashed arrows show the cold path: fetch revision $r$ from shared storage into the per-sampler CPU cache, then admit $L_r$ into a GPU batch slot before inference.}
    \label{fig:mint_system_blocks}
\end{figure}
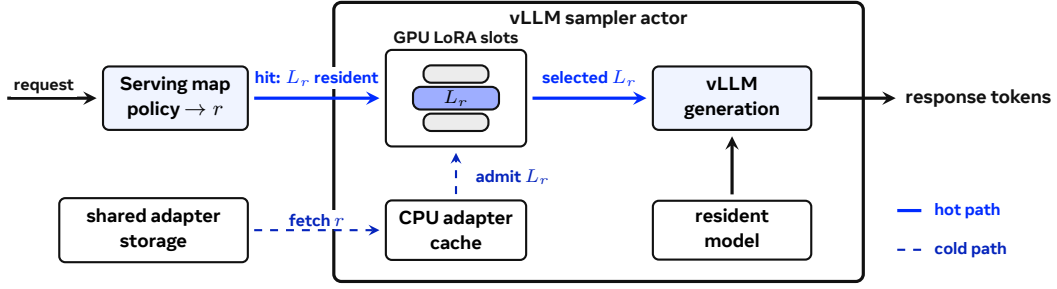

\subsection{Shared-Base Rollout and Serving}

MinT uses vLLM for rollout and serving because vLLM can run inference from a resident base with a LoRA adapter resident in a GPU batch slot. vLLM handles token generation. MinT handles revision selection, adapter load queues, adapter cache tiers, and the export path that turns trainer state into serving adapter files. A sampler actor keeps one base deployment resident, admits exported adapters into vLLM's LoRA slots, and runs inference with the adapter named by MinT. The service maps a user-facing policy name to a policy record, chooses an exported adapter revision, and sends the sampler that revision. \Cref{fig:mint_system_blocks} shows the serving path for a request that resolves to revision $r$.

Each sampler manages adapter cache state across three tiers, which \cref{sec:e4_serving} measures in detail. The addressable catalog holds exported revisions in shared storage. The CPU cache holds adapter bytes already near one sampler actor. The GPU batch holds adapters resident in current inference slots. A request uses a GPU-batch adapter if it is already resident in a slot, promotes a CPU-cached adapter if it is warm, or schedules a cold load from shared storage if the adapter is only in the catalog. Cold adapter loads are scheduled service work before inference: MinT resolves the revision, queues the load job, controls which cold loads enter the sampler, and delays inference starts when too many distinct cold adapters arrive together.

\section{Three Scaling Axes}
\label{sec:scaling}

\Cref{sec:architecture} defined the training-to-serving path: a rollout is generated by a resident base plus an adapter revision, training updates that adapter, and export returns a PEFT adapter file in serving layout. Scaling MinT stresses three parts of that path: large-base placement, exported adapter bytes, and adapter cache state. \textbf{Scale up} keeps LoRA RL usable when dense or MoE bases require model-parallel placement and sparse-routing metadata. \textbf{Scale down} makes the adapter revision, rather than a merged checkpoint, the object that crosses from training to serving. \textbf{Scale out} keeps a large addressable policy catalog selectable while each engine admits only bounded CPU-resident and GPU-active working sets.

\subsection{Scale Up: LoRA RL on Large Dense and MoE Bases}

\paragraph{Megatron placement.}
MinT uses Megatron training groups when the base model is too large for a single PEFT worker. Tensor parallelism shards dense tensors. Expert parallelism assigns MoE experts to rank groups. Dense-module LoRA tensors follow the tensor-parallel shards for the dense weights they modify. Per-expert LoRA tensors are keyed by expert id: the EP shard owns the experts assigned to it, and TP shards the expert matrices inside that owner group when the run also uses tensor parallelism. Shared-expert LoRA is stored once per EP shard and deduplicated during export.

\paragraph{Policy switching.}
The base shards stay resident across policies. A policy switch restores the adapter tensors, optimizer moments, and accumulated gradients on the ranks that will execute the next update. This keeps the policy training state small relative to the resident base while preserving Megatron's sharding rules.

\paragraph{Distributed export.}
Serving needs one PEFT adapter revision that can attach to a resident vLLM base deployment. Export converts the Megatron training view into that serving view: tensor-parallel LoRA slices are gathered, replicated tensors are deduplicated, and expert LoRA tensors are collected from their expert-parallel owners. The exported adapter is the policy revision that later appears in serving, evaluation, rollback, and catalog records.

\paragraph{MoE router replay.}
MoE RL requires training-time scoring to use the expert path that generated each rollout token. R3 identifies router mismatch as a concrete source of instability in MoE RL: small implementation or precision differences can route the same token to different experts during rollout and training~\citep{r3_moe_router2025,chiang2026routerreplay}. R3 records MoE expert routing. For Qwen3-style MoE runs, MinT stores selected expert ids with rollout records. Training can replay a recorded route when the backend can reconstruct that expert path. Training masks a token when the rollout record lacks selected expert ids or the training backend cannot map those ids to its expert-parallel layout.

\paragraph{Sparse-attention provenance.}
Dynamic sparse attention has a separate mismatch channel. In GLM-5 and GLM-5.1, the DSA indexer and top-$k$ path decide which tokens participate in sparse attention; small numerical differences can change that token set~\citep{glm5_2026,stevenchiang2026supportglm5inmint}. MinT removes observed implementation mismatches where the stack exposes a concrete cause: indexer RoPE layout, normalized query/key inputs, deterministic top-$k$ behavior, frozen indexer defaults, long-context THD/CP support, and LoRA loading for DSA target modules. Probability mismatch can remain after those fixes, so MinT uses IcePop-style rollout correction~\citep{ling_every_step2025}: when the training/rollout probability ratio leaves the configured lower--upper trusted band, the token receives zero importance weight. This mitigation filters unsafe scoring terms. It does not replay every DSA indexer choice, and it does not prove that training used the exact sparse-attention token set selected by the inference engine.

\begin{table}[t]
\centering
\scriptsize
\setlength{\tabcolsep}{4pt}
\renewcommand{\arraystretch}{1.20}
\caption{Representative model-family support validated by the current MinT stack.}
\label{tab:supported_model_families}
\fittowidth{%
\begin{tabular}{@{}M{0.16\textwidth}M{0.18\textwidth}M{0.28\textwidth}M{0.32\textwidth}@{}}
\toprule
\apphead Family & Model structure & Scale validated & MinT support path \\
\midrule
\appkey{Qwen3 series} & Dense and MoE variants & 0.6B/4B dense adapters; 30B-A3B and 235B-A22B MoE runs & Single-worker PEFT, Megatron MoE training, expert-route records, adapter export, and shared-base serving \citep{qwen3_2025,chiang2026routerreplay} \\
\addlinespace[3pt]
\appkey{Moonlight \& Kimi K2} & MLA MoE & Moonlight-16B-A3B bring-up; Kimi K2 1.04T countdown-task LoRA RL & MLA/MoE adapter placement, Megatron-Bridge conversion, vLLM serving, and trillion-parameter-class LoRA RL \citep{chiang2026routerreplay,liu2025Build,kimi_k2_2025} \\
\addlinespace[3pt]
\appkey{GLM-5 / GLM-5.1} & MLA, DSA, MTP, and MoE & Frontier agentic family with DSA/MTP training and serving bring-up & DSA/MTP training patches, DSA LoRA target mapping, vLLM custom-forward LoRA loading, bridge conversion, and IcePop rollout correction \citep{glm5_2026,stevenchiang2026supportglm5inmint,ling_every_step2025} \\
\bottomrule
\end{tabular}%
}
\end{table}

\subsection{Scale Down: Adapter-Only Training-Serving Handoff}

\paragraph{Adapter handoff bytes.}
The training-serving handoff uses the LoRA adapter as the checkpoint. A measured Qwen3-4B rank-32 PEFT adapter file is 264{,}310{,}274 bytes, about 252 MiB. Adapter byte size is determined by rank, target modules, dtype, and tensor layout. A four-billion-parameter bf16 base checkpoint has an 8.0 GB weight floor before metadata and optimizer state. The measured adapter is about 3.3\% of that base-weight floor, and the serving path can load it without materializing another full base copy. Since LoRA tensor bytes scale approximately linearly with rank for a fixed target-module set, the same target pattern at rank 1 would be about 7.9 MiB, or roughly 0.10\% of the bf16 base-weight floor. This supports the abstract's conservative 1\% claim: compact rank-1 settings can fall below 1\%, while broader target-module choices and metadata can increase the footprint. The system property remains the same: the crossing artifact is an adapter revision, not a full checkpoint.

\paragraph{Serving-compatible export.}
MoE export applies the same adapter-revision rule to sharded training runs. The trainer writes a PEFT adapter file in the tensor layout expected by vLLM, with optimizer state and rank-local training files removed. The file that crosses from training to serving is the adapter revision.

\paragraph{Served-score attribution.}
A served score names the exported adapter revision together with its compatible base model, target modules, prompt renderer, scorer, and serving path. When a row reports a served score, MinT compares isolated adapter loading against shared-base serving before attributing the served score to the exported revision. This check separates the policy result from loader and routing differences.
\FloatBarrier

\subsection{Scale Out: Policy-Population Serving}
\label{sec:scale_out_residency}

\paragraph{From stored adapters to live policies.}
Scale-out begins when exported adapters stop being a small set of training files and become a large set of deployable policies. The catalog can grow with tenants, product variants, rollback points, personalization branches, evaluation snapshots, and research sweeps. A user-facing name resolves to an exported adapter revision, the revision resolves to a durable adapter file, and a serving actor maps that file to a local adapter id when the policy enters that actor. \Cref{tab:scale_out_cache_tiers} names the three tiers that a policy revision can occupy at once; the paragraphs below visit them in turn.

\begin{table}[t]
\centering
\scriptsize
\setlength{\tabcolsep}{3.2pt}
\renewcommand{\arraystretch}{1.10}
\caption{Three cache tiers separate addressability from local residency and same-batch execution. A policy revision can occupy any subset of these tiers at a given moment; the scales, lifetimes, and control surfaces differ.}
\label{tab:scale_out_cache_tiers}
\fittowidth{%
\begin{tabular}{@{}M{0.16\textwidth}M{0.15\textwidth}M{0.15\textwidth}M{0.22\textwidth}M{0.26\textwidth}@{}}
\toprule
\apphead Tier & Scale & Lifetime & Promotion / eviction & Measured in this paper \\
\midrule
\appkey{Addressable catalog} & $10^3$--$10^6$ entries & Durable (control plane) & Promoted by adapter export; retired manually & Built and audited 1M packed catalog (Appendix~\cref{tab:app_1m_catalog_audit}); selected-revision serving (\cref{tab:e4_serving_summary}) \\
\addlinespace[3pt]
\appkey{CPU adapter cache} & Hundreds per engine & Per actor run & Promoted by router or cache-miss load; LRU under memory pressure & 369 / 550 cached on one engine (\cref{tab:e4_serving_summary}; Appendix~\cref{tab:app_cache_ladders}) \\
\addlinespace[3pt]
\appkey{GPU batch} & $\le 64$ distinct adapters & One decoding step & Promoted by the batch scheduler; released at end of step & Same-batch frontier $N{=}64$ (\cref{tab:e4_serving_summary}; Appendix~\cref{tab:app_cache_ladders}) \\
\bottomrule
\end{tabular}%
}
\end{table}

\paragraph{Catalog size is the request-name scale.}
The 1M packed-catalog experiment keeps name scale separate from engine-local cache state. MinT materializes 1{,}000{,}000 packed adapter revisions in 100 shards and audits 256 sampled adapters across all shards, so the million-scale catalog is a measured artifact rather than a sizing extrapolation. The Qwen3-30B TP=4 serving experiments then select bounded working sets from that catalog: one actor keeps hundreds of adapters cached near the engine and runs clean same-batch probes through 64 distinct adapters. These measurements split scale-out into three quantities: how many policy revisions can be named, how many can stay cached near an engine, and how many can execute in the current batch. The $10^6$ number is addressability over a built catalog, not simultaneous GPU residency.

\paragraph{Traffic locality drives cache state.}
Catalog membership and local cache state change on different time scales. Catalog entries live in the control plane. Cache entries are local to a serving actor. The same adapter revision can be addressable in the catalog, cached in CPU memory on one actor, active in the current GPU batch, evicted back to shared storage, and later loaded again. Recurring traffic should stay near an engine that already holds the selected policy, while weak-locality sweeps and rollout waves expose cache misses.

\paragraph{Cold loading is service work.}
A cache miss does more than read a small LoRA file. The serving actor fetches tensors, materializes loader objects, registers the adapter with the engine, and activates it before decoding can start. Different missing policies serialize through the engine load path, so latency grows as a staircase when many unique policies arrive together. Concurrent cache-miss requests for the same missing policy can share one load; requests for different missing policies remain separate load jobs. MinT therefore treats cold loading as scheduled service work with deduplication and bounded backpressure.

\paragraph{Serving contract.}
Policy-population serving requires controls after name lookup. Routing preserves CPU-cache reuse when traffic has locality. Batch construction respects the smaller same-batch adapter limit; the addressable catalog is larger than any realized batch. Cold loading is observable, retryable, and backpressured because it turns a stored adapter revision into an engine-local adapter object. MinT also separates \emph{registered}, \emph{prewarming}, and \emph{ready}: online catalog registration can make a revision known to the control plane, while the serving plane exposes it to user traffic only after the readiness/prewarm phase activates the adapter under admission control. The ready state therefore means that user requests no longer perform adapter loading; it does not mean that registration and activation are instantaneous. The policy record and exported revision stay stable while the adapter moves among shared storage, the CPU cache, the GPU batch, and readiness states.

\paragraph{Representation controls the load staircase.}
MoE LoRA makes the cold path visible because a moderate-size adapter can expand into tens of thousands of tiny tensor objects and hundreds of megabytes of cached state across TP workers. MinT packs these tensors into a serving representation with nearly unchanged declared bytes, reducing object fanout before engine registration. The detailed serving measurements in \cref{sec:e4_serving} show 37{,}248 tensors reduced to 672 and an 8.5--8.7$\times$ live-load speedup, with the packed live engine-load slice below 0.2 seconds in the measured bursts. This live-load number is one component of a longer cold path that also includes routing, queueing, fetch, activation, readiness waiting, and generation. The mixed warm/cold reload measurements show why the stage must be scheduled: cold first-touch can push old warm-tenant TTFT p95 above 20 seconds, while admission-aware two-phase rollout keeps warm stalls bounded and removes adapter-loading time from user-visible first requests after readiness. The exported adapter remains the policy unit, while cold loading becomes an explicit adapter lifecycle stage that can be routed, prewarmed, throttled, and optimized.

\section{Evaluation}
\label{sec:evaluation}

The experiments test the three scaling axes through the same LoRA lifecycle. Scale Down is measured by adapter-only handoff against merge-and-load paths and by concurrent training schedules that reuse one resident base allocation. Scale Up is measured by dense SFT, DPO, and GRPO runs, sparse-route MoE RL, Qwen3-235B-A22B GRPO, GLM-5.1 A2UI GRPO, and a Kimi K2 1T countdown-task path. Scale Out is measured by separating a materialized 1M packed adapter catalog from CPU-cache size, same-batch adapter diversity, cold-load cost, and online rollout control. The million-scale claim is an artifact-backed catalog claim plus selected-revision serving measurements, not a claim that one engine keeps every adapter resident or active.

An adapter revision means a fixed exported adapter version selected by rollout, evaluation, serving, or rollback. The term names the behavior selected by a request, separate from the file transfer or load operation that may place that adapter near a sampler.

The handoff and utilization measurements isolate the systems effect of making the adapter the training-serving artifact. The learning measurements confirm that the same lifecycle supports the public cookbook recipes and larger sparse-model deployments. The serving measurements show that a large policy catalog is an addressability scale, while CPU-cache residency, same-batch adapter diversity, cold-load work, and readiness state are the online execution scales. This distinction is the evidence boundary for the abstract: MinT can manage a measured million-scale catalog while training and serving selected revisions through bounded resident working sets.

SFT denotes supervised fine-tuning. DPO denotes pairwise preference optimization. GRPO denotes rollout-based reinforcement learning. The public MinT cookbook is the recipe layer around the framework: it packages task configurations, benchmark manifests, proxy screens, full confirmations, and maintained adapter recipes~\citep{mint_cookbook2026}. DAPO-AIME24 is the math-RL cookbook recipe evaluated on AIME 2024, chat-DPO is the pairwise-preference recipe, LawBench is the legal-reasoning recipe, and Fineval is a finance-domain supervised benchmark.

\subsection{Scale Down and Multi-Train Utilization}
\label{sec:e12_pilots}

\begin{figure}[t]
    \centering
    \includegraphics[width=0.92\textwidth]{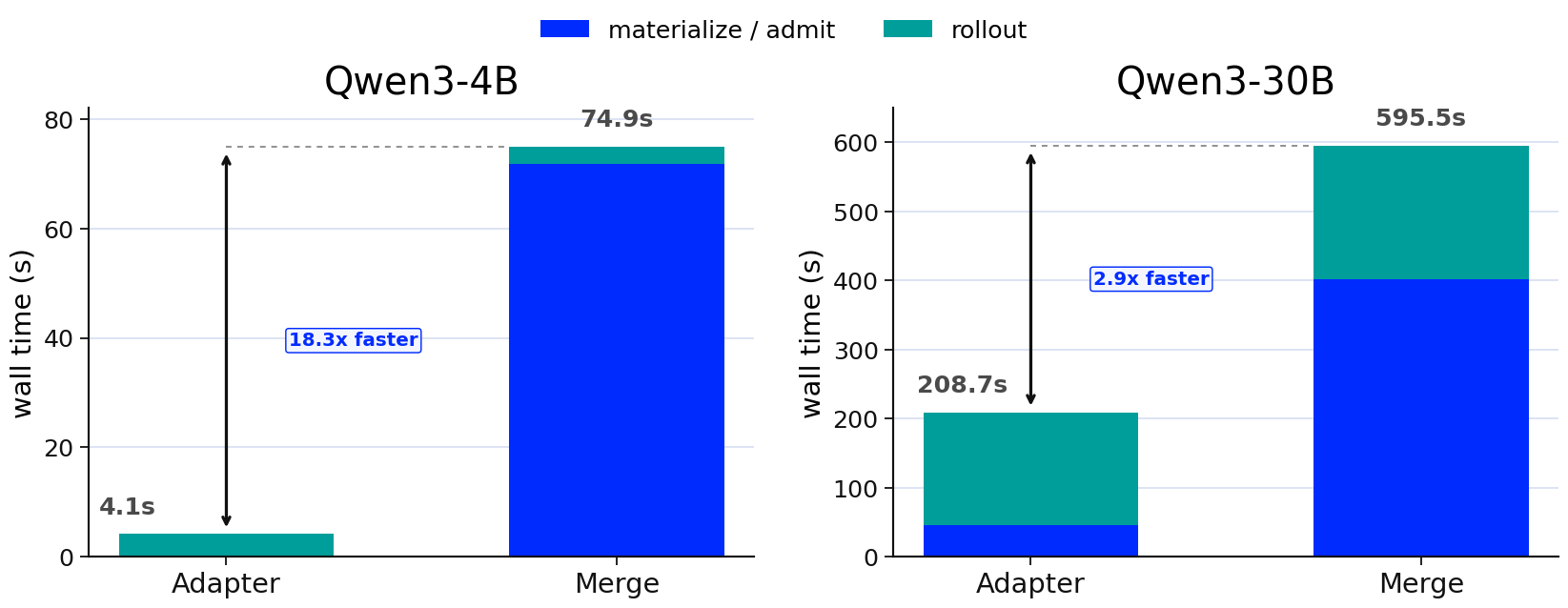}
    \caption{Adapter handoff avoids the merge-and-load stages that dominate the full-checkpoint path. Each stacked bar separates materialization or adapter loading from rollout under the probe protocol for the corresponding model.}
    \label{fig:e1_handoff_breakdown}
\end{figure}

Adapter handoff compares two ways to send a newly trained policy to the sampler. The merge path materializes a full checkpoint and then loads that checkpoint before rollout. The MinT path loads the exported adapter into a resident shared-base sampler. \Cref{fig:e1_handoff_breakdown} plots total step time as materialization or adapter loading plus rollout, and \cref{tab:e1_handoff_paths} records the file sizes, cold first-sample latency, and total versus warm generation rates used to interpret the bars. The total rate includes the first request in the probe sequence, while the warm rate excludes it. A merged checkpoint may achieve higher or lower token throughput than a LoRA-based adapter during rollout, so the end-to-end comparison involves a trade-off between the cost of shipping the artifact and the resulting sampling throughput. In these runs, loading the adapter saves enough materialization and loading time to dominate the rollout-speed differences.

\begin{table}[H]
\centering
\scriptsize
\setlength{\tabcolsep}{3pt}
\renewcommand{\arraystretch}{1.05}
\caption{File and sampling metrics for the adapter-handoff comparison in \cref{fig:e1_handoff_breakdown}. Cold first sample is the first request wall time. The total sample speed includes that first request, while the warm sample speed excludes it.}
\label{tab:e1_handoff_paths}
\fittowidth{%
\begin{tabular}{@{}M{0.11\textwidth}M{0.12\textwidth}M{0.16\textwidth}M{0.14\textwidth}M{0.17\textwidth}M{0.13\textwidth}M{0.15\textwidth}@{}}
\toprule
\apphead Model & Path & Checkpoint parameters & Checkpoint file size & Materialization or load & Cold first sample & \makecell{sample speed\\total/warm} \\
\midrule
\appgroup{7}{Qwen3-4B}
Qwen3-4B & \appkey{Adapter} & rank-32 LoRA & 252 MiB & 0.036 s & 4.114 s & 15.568/15.567 tok/s \\
Qwen3-4B & \appkey{Merge} & full model & 8.061 GB & 71.820 s & 55.704 s & 4.697/20.595 tok/s \\
\appgroup{7}{Qwen3-30B}
Qwen3-30B & \appkey{Adapter} & rank-16 LoRA & 1.692 GB & 46.455 s & 117.304 s & 1.874/5.700 tok/s \\
Qwen3-30B & \appkey{Merge} & full model & 61.084 GB & 402.245 s & 156.074 s & 1.573/6.904 tok/s \\
\bottomrule
\end{tabular}%
}
\end{table}

\begin{figure}[H]
    \centering
    \includegraphics[width=\textwidth]{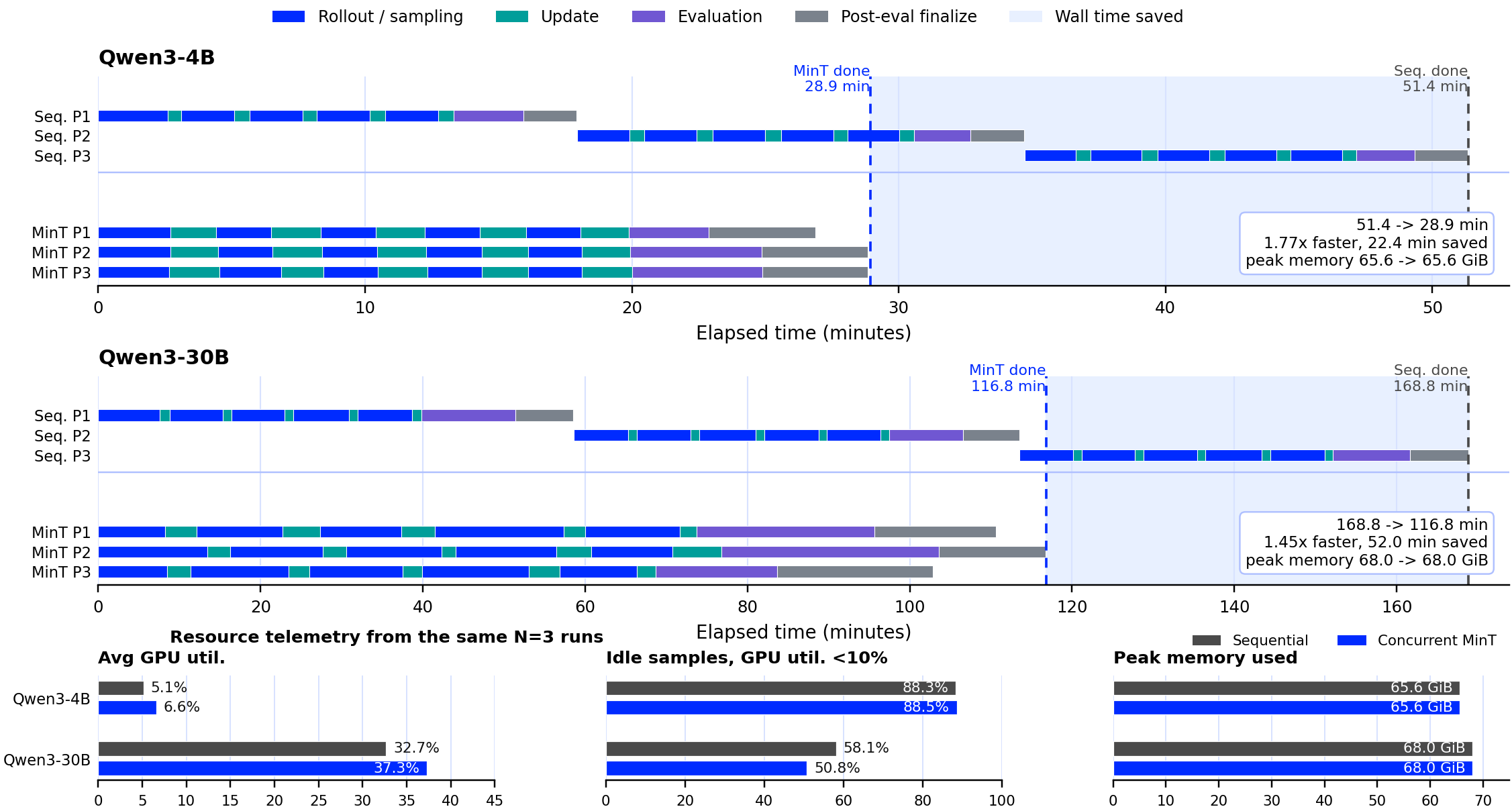}
    \caption{Concurrent multi-LoRA training overlaps GRPO runs under the same base-model allocation. Timeline lanes show the schedule, and the lower panels summarize average GPU utilization, samples below 10\% utilization, and peak memory from the same runs. Vertical dashed lines mark completion time for the concurrent and sequential schedules.}
    \label{fig:e2_gpu_utilization}
\end{figure}

Concurrent training measures schedule utilization under one resident base allocation. A sequential schedule can fit in memory while leaving the resident base idle between rollout, update, export, and evaluation phases. MinT keeps the same peak memory budget and lets other policies use those idle periods on resident trainers and samplers. The measured speedup comes from filling cross-policy gaps in the schedule, while each rollout and optimizer step still runs the same model computation for its selected policy. In the completed GRPO runs, concurrent execution finishes in 1736.1 s instead of 3081.2 s on Qwen3-4B and in 7008.4 s instead of 10130.0 s on Qwen3-30B, while peak memory remains unchanged within each model size. \Cref{fig:e2_gpu_utilization} shows the measured schedule timeline and matching GPU telemetry, and \cref{tab:e2_training_utilization} records the schedule summaries. The figure/table pair is therefore a systems-utilization check: the timeline shows idle gaps being filled, and the table reports the wall-clock effect under the same peak-memory envelope.

\begin{table}[H]
\centering
\scriptsize
\setlength{\tabcolsep}{3.2pt}
\renewcommand{\arraystretch}{1.10}
\caption{Concurrent-training schedule summary for three GRPO policies. Time saved and speedup compare each concurrent MinT row against the sequential row for the same model.}
\label{tab:e2_training_utilization}
\fittowidth{%
\begin{tabular}{@{}M{0.14\textwidth}M{0.19\textwidth}M{0.16\textwidth}M{0.15\textwidth}M{0.19\textwidth}M{0.10\textwidth}M{0.13\textwidth}@{}}
\toprule
\apphead Model & Schedule & Workloads & Wall time & Time saved & Speedup & Peak memory \\
\midrule
\appgroup{7}{Qwen3-4B}
Qwen3-4B & \appkey{Sequential} & 3 GRPO policies & 3081.2 s & -- & $1.00\times$ & 65.6 GiB \\
Qwen3-4B & \appkey{Concurrent MinT} & 3 GRPO policies & 1736.1 s & 1345.1 s / 43.7\% & $1.77\times$ & 65.6 GiB \\
\appgroup{7}{Qwen3-30B}
Qwen3-30B & \appkey{Sequential} & 3 GRPO policies & 10130.0 s & -- & $1.00\times$ & 68.0 GiB \\
Qwen3-30B & \appkey{Concurrent MinT} & 3 GRPO policies & 7008.4 s & 3121.6 s / 30.8\% & $1.45\times$ & 68.0 GiB \\
\bottomrule
\end{tabular}%
}
\end{table}

The timing and utilization experiments isolate the systems effect of the adapter design. Learning quality is evaluated under the same adapter lifecycle in the next group of experiments.
\FloatBarrier

\subsection{Scale Up Across Training Paradigms and Model Scales}
\label{sec:e3_curves}

\begin{figure}[t]
    \centering
    \includegraphics[width=\textwidth]{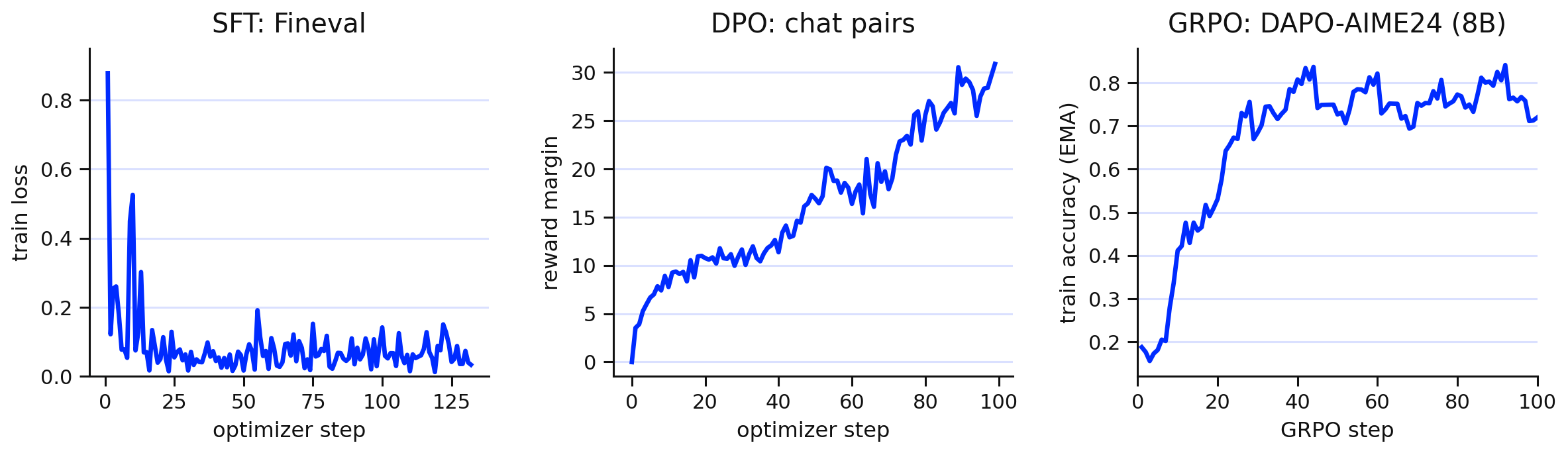}
    \caption{Dense-model learning traces. Each panel keeps the native metric for its training paradigm instead of forcing SFT loss, DPO reward margin, and GRPO train accuracy onto one axis.}
    \label{fig:e3_dense_curves}
\end{figure}

Dense-model experiments use Qwen3-4B for SFT and DPO, and Qwen3-8B base for GRPO. The SFT rows use FinEval and finance-sentiment benchmarks from the FinGPT evaluation suite~\citep{fineval2023,fingpt2023}. The DPO row uses the chat-DPO recipe's reward-margin trace. The GRPO rows use the Qwen3-8B-base DAPO-AIME24 trace on the 2024 American Invitational Mathematics Examination~\citep{maa_aime2024}. \Cref{fig:e3_dense_curves} shows the learning traces, while \cref{tab:e3_dense_results} separates curve-backed rows from held-out confirmation rows.

Each paradigm exercises a different part of the same lifecycle. SFT tests whether a supervised dataset moves into the adapter, trains through the LoRA path, and produces held-out gains comparable to a full fine-tune; the five FinGPT-suite rows cover finance-domain accuracy and the broader FinEval task at the same time. DPO tests whether a preference-pair objective drives the same adapter through the chat-DPO recipe and increases the chosen-minus-rejected reward margin during optimization. GRPO tests whether a rollout-based RL recipe updates the adapter from on-policy AIME 2024 rollouts. We do not claim that the three benchmarks are comparable as scores; we claim that one adapter type, one export format, and one serving path carry SFT loss, DPO reward margin, and GRPO train accuracy without any per-paradigm tooling change. \Cref{tab:e3_dense_results} reports the endpoint rows that quantify each of these checks.

The result rows in \cref{tab:e3_dense_results} separate two kinds of evidence under the same adapter lifecycle. The five SFT rows from the FinGPT evaluation suite all confirm large held-out gains over the base Qwen3-4B score: roughly $+36$ points on FinEval, $+19$ on FPB, $+31$ on TFNS, and smaller but consistent moves on FiQA-SA and NWGI. The DPO row reports the reward-margin endpoint from the chat-DPO optimization trace. The GRPO rows report the Qwen3-8B-base DAPO-AIME rollout, update, export, and evaluation path: the train-side accuracy EMA rises over the plotted run segment, and the exported step-240 sampler solves 20/30, 19/30, and 18/30 problems on the AIME24, AIME25, and AIME26 eval sets with temperature 1.0, top-p 0.7, a 32K generation cap, and seed 42. The combined message is paradigm-agnostic: the same lifecycle carries supervised, preference-based, and rollout-based updates without any per-paradigm checkpoint surgery.

\begin{table}[t]
\centering
\scriptsize
\setlength{\tabcolsep}{3.2pt}
\renewcommand{\arraystretch}{1.10}
\caption{Dense-model result rows. The table keeps the native metric for each objective and separates plotted traces from held-out confirmation scores. FPB, FiQA-SA, TFNS, and NWGI are finance-sentiment held-out sets from the FinGPT evaluation suite~\citep{fingpt2023}.}
\label{tab:e3_dense_results}
\fittowidth{%
\begin{tabular}{@{}M{0.14\textwidth}M{0.25\textwidth}M{0.16\textwidth}M{0.17\textwidth}M{0.20\textwidth}@{}}
\toprule
\apphead Paradigm & Benchmark & Metric & Evidence & Result \\
\midrule
\appgroup{5}{SFT: finance suite}
\appkey{SFT} & Fineval & accuracy & held-out score & 0.4226 $\rightarrow$ 0.7811 \\
\appkey{SFT} & FPB & accuracy & held-out score & 0.6906 $\rightarrow$ 0.8804 \\
\appkey{SFT} & FiQA-SA & accuracy & held-out score & 0.8255 $\rightarrow$ 0.8473 \\
\appkey{SFT} & TFNS & accuracy & held-out score & 0.5959 $\rightarrow$ 0.9095 \\
\appkey{SFT} & NWGI & accuracy & held-out score & 0.4954 $\rightarrow$ 0.5925 \\
\appgroup{5}{Preference optimization}
\appkey{DPO} & chat pairs & reward margin & trace endpoint & $-0.03 \rightarrow 30.88$ \\
\appgroup{5}{Reinforcement learning}
\appkey{GRPO} & AIME24 & train accuracy (EMA) & Qwen3-8B trace & 0.188 $\rightarrow$ 0.719; peak 0.841 at step 92 \\
\appkey{GRPO} & AIME24/25/26 & eval accuracy & step-240 sampler & 20/30, 19/30, 18/30 \\
\bottomrule
\end{tabular}%
}
\end{table}

\begin{figure}[t]
    \centering
    \includegraphics[width=0.98\textwidth]{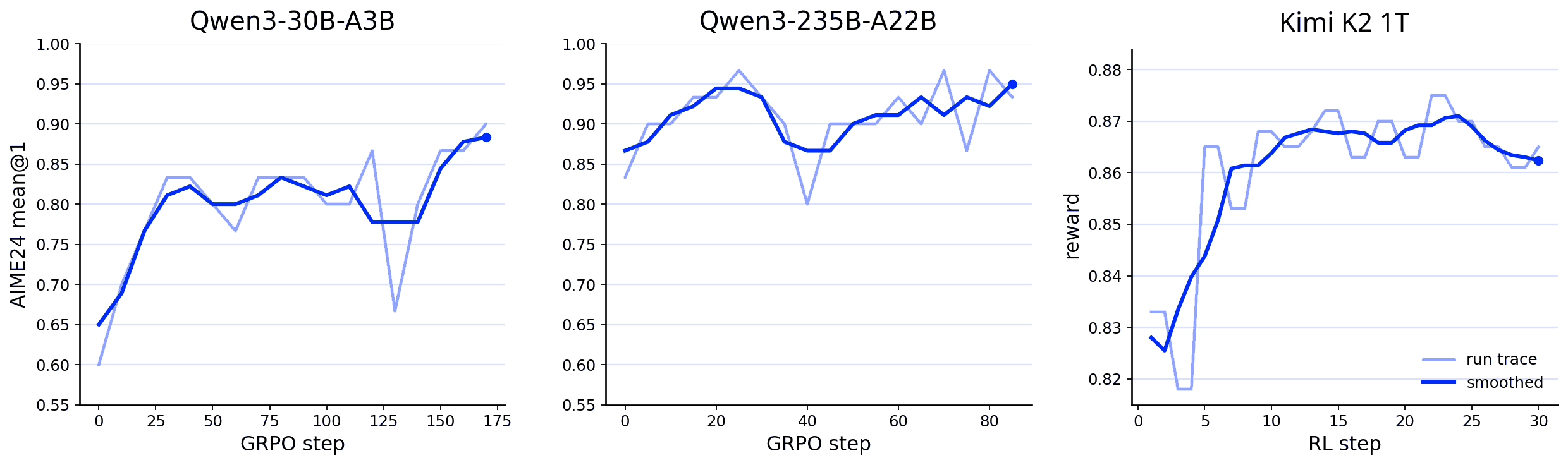}
    \caption{MoE RL curves. The 30B/235B panels use AIME24 mean@1 with aligned y axes and smoothed overlays. The Kimi K2 panel gives the end-to-end LoRA RL reward curve for the 1T countdown-task run~\citep{liu2025Build}.}
    \vspace{-10pt}
    \label{fig:e3_moe_curves}
\end{figure}

MoE experiments add two conditions beyond the dense rows. First, expert routes must be replayed during training-time scoring so that tokens are scored under the MoE path that generated them. Second, large MoE models test whether the adapter/base split survives distributed placement across tensor-parallel and expert-parallel workers. The Qwen3-235B-A22B run uses a 32-A800-GPU Megatron trainer with TP=4 and EP=8 (PP=1), paired with a 16-GPU TP=16 serving deployment. The Kimi K2 1.04T countdown-task run uses a 64-GPU H800 deployment with the same LoRA RL path on 32.6B active parameters. \Cref{fig:e3_moe_curves} shows the 30B and 235B AIME24 curves together with the Kimi K2 1T countdown-task RL curve~\citep{liu2025Build}.

The 30B-A3B and 235B-A22B panels share the same y axis to make the scale jump readable. The 30B-A3B AIME24 curve rises from a noisy near-zero start to a stable mid-band by the end of the logged window, while the 235B-A22B curve reaches 0.967 peak mean@1 -- close to saturation on AIME24 under the same LoRA RL path. The Kimi K2 panel switches to task reward instead of an AIME-style accuracy because the countdown task has a different correctness target, but the curve follows the same rollout-update-export-evaluate loop on a 1.04T-parameter base. Together, the three panels close the scale-up claim along the adapter lifecycle: the LoRA adapter remains the policy object across a 30B sparse base, a 235B-A22B distributed deployment, and a trillion-parameter MoE, with no change to the training-serving handoff between them.

\begin{figure}[t]
    \centering
    \begin{subfigure}[t]{0.48\textwidth}
        \centering
        \includegraphics[width=\linewidth]{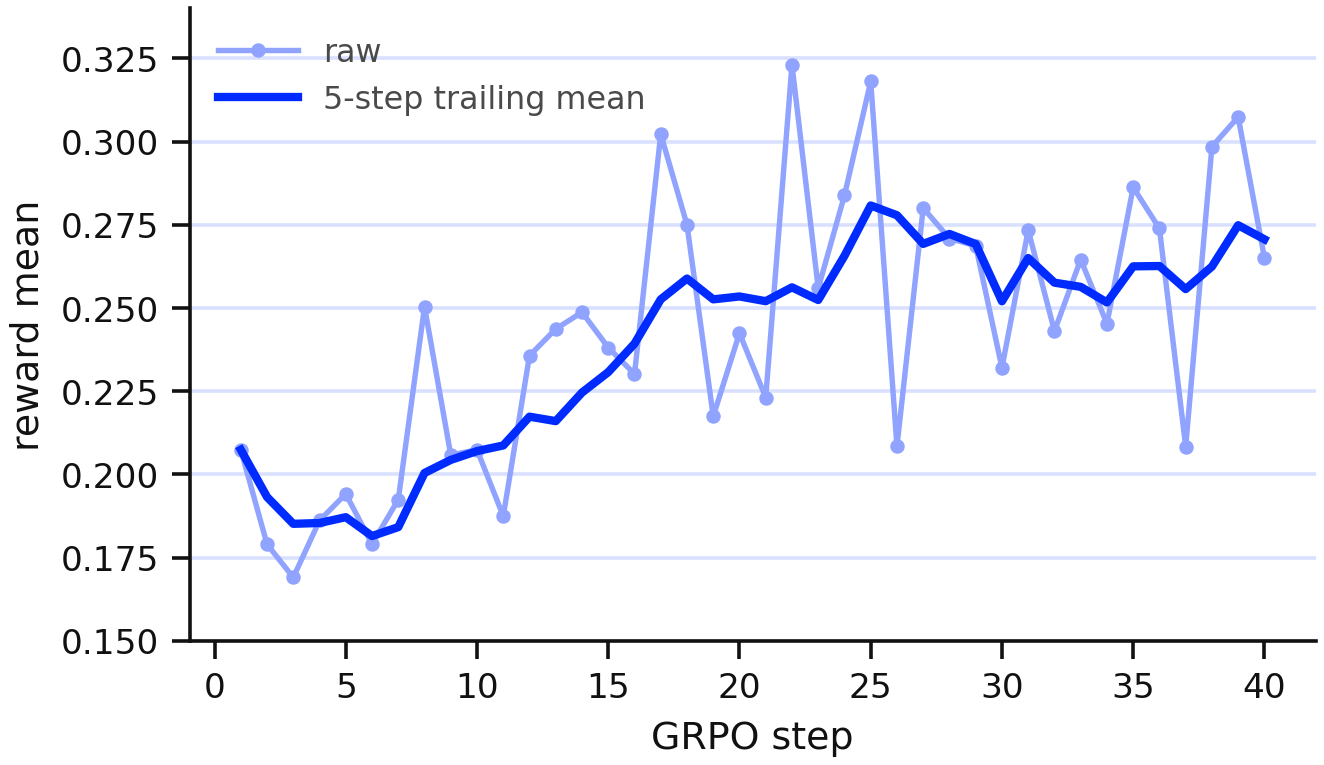}
        \caption{Reward}
    \end{subfigure}
    \hfill
    \begin{subfigure}[t]{0.48\textwidth}
        \centering
        \includegraphics[width=\linewidth]{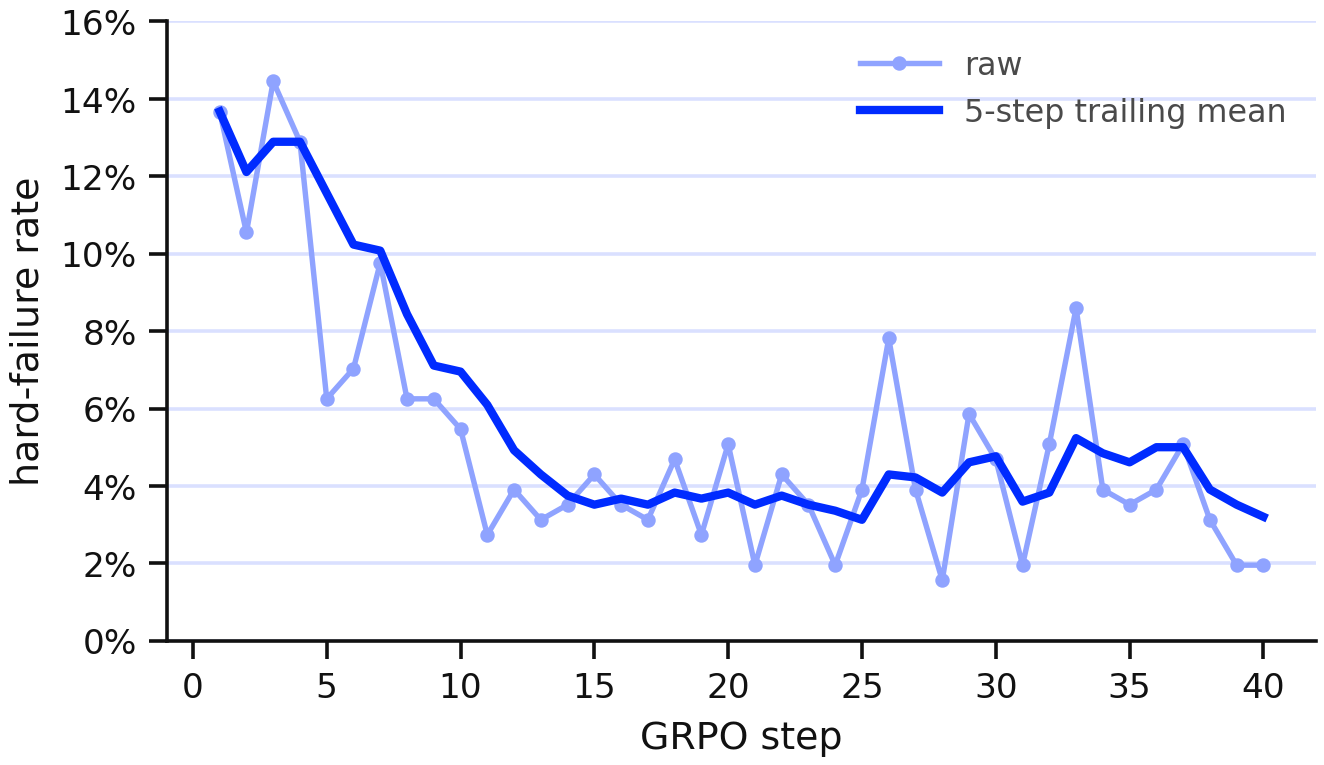}
        \caption{Hard failures}
    \end{subfigure}
    \caption{GLM-5.1 A2UI GRPO curves. Each step samples 32 prompts with 8 responses per prompt; raw points report the per-step mean over 256 responses, and thick curves show a 5-step trailing mean. Reward uses the A2UI reward stack. Hard failures are responses assigned a hard-fail reward note; by step 40 the run records 5 hard failures out of 256 responses.}
    \label{fig:e3_glm51_a2ui_curves}
\end{figure}

Generative UI gives MinT a product-facing scale-up case beyond math benchmarks. We train a GLM-5.1 LoRA policy for Macaron-A2UI, a state-of-the-art Generative UI model where the model must emit natural language together with executable A2UI actions for a trusted client renderer~\citep{macaron_a2ui2026}. The GRPO run resumes across multiple training jobs and keeps a fixed sampling contract: 32 prompts per step, 8 responses per prompt, and a reward stack that separates executable interaction quality from contract-level hard failures. Over 40 GRPO steps, the 5-step reward mean rises from about 0.21 to 0.27, while the hard-failure rate drops from the low teens early in training to 5/256 responses at step 40. This run exercises the same MinT adapter lifecycle on a frontier Generative UI workload: rollout sampling, reward computation, LoRA update, checkpoint export, and serving re-entry happen without moving a full GLM-5.1 model copy between stages.

\begin{figure}[t]
    \centering
    \includegraphics[width=\textwidth]{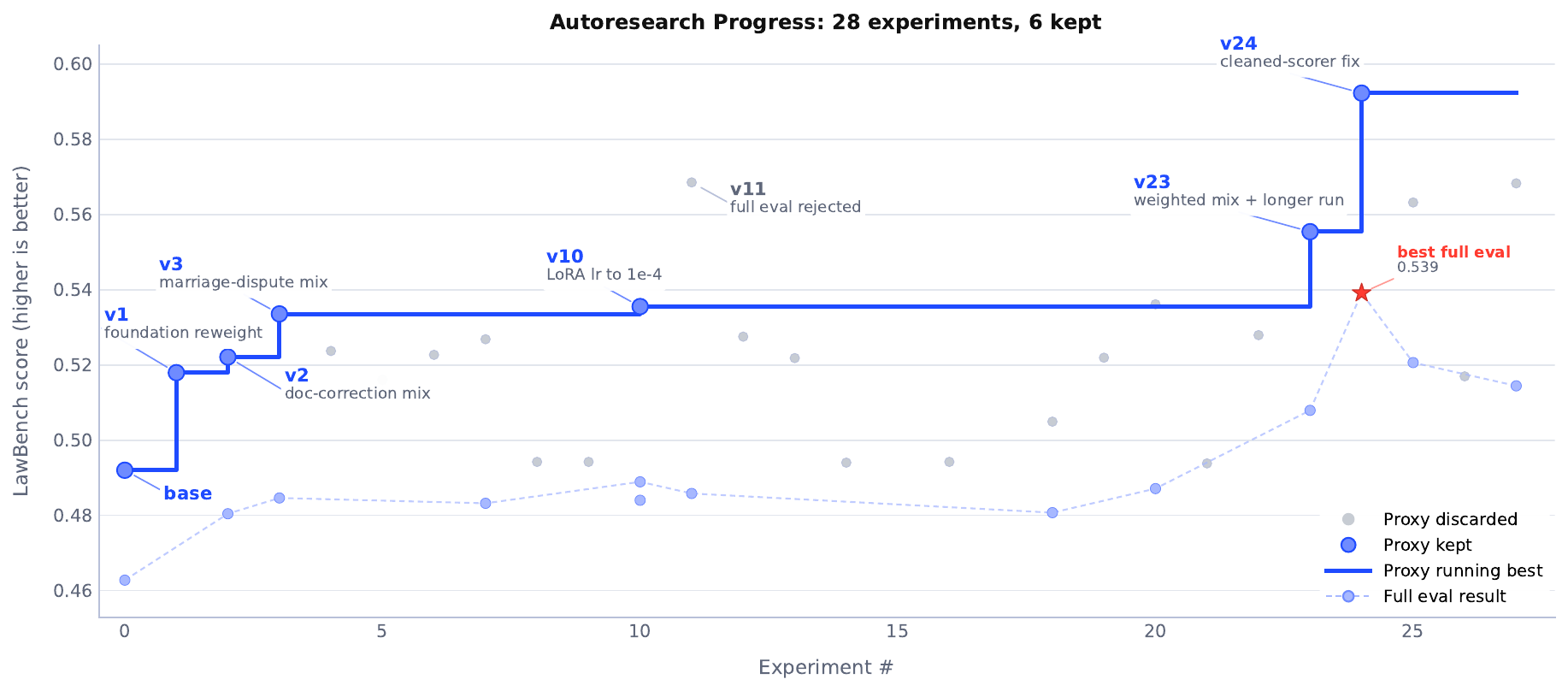}
    \caption{LawBench AutoResearch trace from the cookbook utilities. Pale gray points are proxy-screened candidates that were not promoted, blue-outlined points are kept proxy candidates, the blue step line is the running best among kept candidates, violet points are full LawBench evaluations, and the black diamond is the full-manifest control. The labeled v11 proxy high was rejected after full-benchmark confirmation.}
    \label{fig:e3_autoresearch_lawbench}
\end{figure}

The MoE route counters report how often token-level scoring attempts needed route-consistency handling. On Qwen3-30B, the run with R3 has a mean out-of-route scoring ratio of 0.0013\% across 87 logged steps, while the comparable no-R3 run averages 0.0097\% across 50 logged steps. On Qwen3-235B, the R3 run averages 0.0253\% across 88 logged steps. These counters are small because they are token-level rates, yet they identify the exact failure channel that R3 controls: a token sampled under one expert route should not be scored as if a different route had produced it. The Qwen3-235B curve reaches 0.967 peak mean@1 on AIME24.

AutoResearch evaluates recipe candidates in two stages. Proxy LawBench tasks screen many data-manifest and recipe variants, and selected variants then receive full LawBench confirmation before being counted as improved recipes~\citep{lawbench2024}. FinGPT, DAPO-AIME, and chat-DPO supply confirmed SFT, GRPO, and DPO rows above. The cookbook provides AutoResearch utilities for coding agents to launch MinT runs, screen candidate recipes on proxy LawBench tasks, and promote selected candidates to full LawBench confirmation. In the LawBench run, the base Qwen3-4B full score is 0.4628. The v10 recipe is the earlier learning-rate-tuned line. The high-proxy v11 candidate remains discarded because its full LawBench score is 0.4858, below v10's 0.4889. The full-manifest control reruns the later data manifest without the promoted changes and reaches 0.4712. The maintained v23 weighted-aligned recipe extends the aligned-data line to a fairer exposure budget; it reaches 0.5554 on the proxy screen at step 100 and 0.5079 on the full benchmark~\citep{mint_cookbook2026}. \Cref{fig:e3_autoresearch_lawbench} plots the proxy-search trajectory and full-benchmark confirmation points.

The proxy-versus-full divergence is the methodological signal the cookbook is built around. The v11 candidate set a new proxy high but its full LawBench score fell below v10's; this is the exact failure mode an unsupervised proxy can produce when its task distribution diverges from the full benchmark's. The full-manifest control at 0.4712 shows that the later data manifest alone does not produce the gain, which separates data effects from recipe effects in the same trace. The maintained v23 weighted-aligned line is the candidate that survives both stages, with a proxy reading of 0.5554 at step 100 and a full reading of 0.5079, well above the base 0.4628 and the v10/v11 cluster. Reading the running-best line together with the kept-candidate filter is how AutoResearch decides to record progress instead of celebrating a single high point, and it is the same lifecycle pattern the previous paradigms use: train an adapter, freeze it as a revision, evaluate that exact revision, then decide whether it enters the maintained recipe set.

\FloatBarrier

\subsection{Policy-Population Serving}
\label{sec:e4_serving}

Multi-LoRA serving becomes a policy-population problem when many exported adapter revisions are deployable at once. Training still exports a LoRA adapter instead of a full checkpoint, and serving still keeps the base model resident. A request names one policy revision; the serving actor either finds the adapter in the GPU batch, promotes it from the CPU cache, or loads it from shared storage after a cache miss. Only the policies selected by the current batch consume GPU-batch adapter slots.

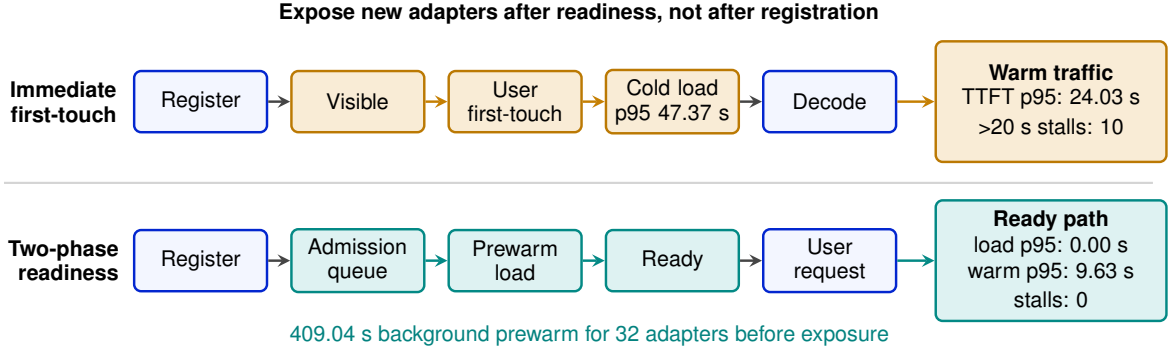
\begin{figure}[t]
    \centering
    \resizebox{0.98\textwidth}{!}{\begin{tikzpicture}[x=1cm,y=1cm,>=stealth,font=\fontfamily{phv}\selectfont]
  \definecolor{mintauxteal}{HTML}{009E9A}
  \definecolor{mintauxamber}{HTML}{D98B00}
  \tikzstyle{lane}=[font=\scriptsize\fontfamily{phv}\selectfont\bfseries, align=right]
  \tikzstyle{box}=[draw=mindlabfg, thick, rounded corners=0.8mm, fill=white,
                   minimum height=0.70cm, inner xsep=2pt, inner ysep=3pt,
                   font=\scriptsize\fontfamily{phv}\selectfont, align=center]
  \tikzstyle{procbox}=[box, text width=1.38cm, minimum width=1.52cm]
  \tikzstyle{badstep}=[procbox, fill=mintauxamber!16, draw=mintauxamber!85!mindlabfg]
  \tikzstyle{goodstep}=[procbox, fill=mintauxteal!13, draw=mintauxteal!85!mindlabfg]
  \tikzstyle{plainstep}=[procbox, fill=mindlabbluepale!35, draw=mindlabblue!80!mindlabfg]
  \tikzstyle{outcomebad}=[box, fill=mintauxamber!16, draw=mintauxamber!85!mindlabfg,
                          text width=2.50cm, minimum height=1.38cm]
  \tikzstyle{outcomegood}=[box, fill=mintauxteal!13, draw=mintauxteal!85!mindlabfg,
                           text width=2.50cm, minimum height=1.38cm]
  \tikzstyle{outtitle}=[font=\scriptsize\fontfamily{phv}\selectfont\bfseries, align=center]
  \tikzstyle{outmetric}=[font=\scriptsize\fontfamily{phv}\selectfont, align=center]
  \tikzstyle{arrow}=[->, thick, draw=mindlabfg!78]
  \tikzstyle{badarrow}=[->, thick, draw=mintauxamber!90!mindlabfg]
  \tikzstyle{goodarrow}=[->, thick, draw=mintauxteal!85!mindlabfg]
  \tikzstyle{note}=[font=\scriptsize\fontfamily{phv}\selectfont, align=center]

  \path[use as bounding box] (0.00,-0.15) rectangle (13.95,4.45);

  \node[note, font=\scriptsize\fontfamily{phv}\selectfont\bfseries] at (6.98,4.17)
    {Expose new adapters after readiness, not after registration};

  \node[lane] at (1.08,3.16) {Immediate\\first-touch};
  \node[plainstep] (regA) at (2.65,3.16) {Register};
  \node[badstep] (visA) at (4.45,3.16) {Visible};
  \node[badstep] (touchA) at (6.25,3.16) {User\\first-touch};
  \node[badstep] (loadA) at (8.05,3.16) {Cold load\\p95 47.37 s};
  \node[plainstep] (decodeA) at (9.85,3.16) {Decode};
  \node[outcomebad, anchor=west] (badout) at (11.05,3.16) {};
  \node[outtitle] at ([yshift=0.34cm]badout.center) {Warm traffic};
  \node[outmetric] at ([yshift=0.03cm]badout.center) {TTFT p95: 24.03 s};
  \node[outmetric] at ([yshift=-0.28cm]badout.center) {\textgreater{}20 s stalls: 10};
  \draw[arrow] (regA) -- (visA);
  \draw[badarrow] (visA) -- (touchA);
  \draw[badarrow] (touchA) -- (loadA);
  \draw[arrow] (loadA) -- (decodeA);
  \draw[badarrow] (decodeA) -- (badout.west);

  \node[lane] at (1.08,1.34) {Two-phase\\readiness};
  \node[plainstep] (regB) at (2.65,1.34) {Register};
  \node[goodstep] (admB) at (4.45,1.34) {Admission\\queue};
  \node[goodstep] (preB) at (6.25,1.34) {Prewarm\\load};
  \node[goodstep] (readyB) at (8.05,1.34) {Ready};
  \node[plainstep] (serveB) at (9.85,1.34) {User\\request};
  \node[outcomegood, anchor=west] (goodout) at (11.05,1.34) {};
  \node[outtitle] at ([yshift=0.46cm]goodout.center) {Ready path};
  \node[outmetric] at ([yshift=0.16cm]goodout.center) {load p95: 0.00 s};
  \node[outmetric] at ([yshift=-0.14cm]goodout.center) {warm p95: 9.63 s};
  \node[outmetric] at ([yshift=-0.44cm]goodout.center) {stalls: 0};
  \draw[arrow] (regB) -- (admB);
  \draw[goodarrow] (admB) -- (preB);
  \draw[goodarrow] (preB) -- (readyB);
  \draw[arrow] (readyB) -- (serveB);
  \draw[goodarrow] (serveB) -- (goodout.west);

  \node[note, text=mintauxteal!80!mindlabfg] at (7.10,0.48)
    {409.04 s background prewarm for 32 adapters before exposure};

  \draw[mindlabfg!18, line width=0.7pt] (0.40,2.23) -- (13.55,2.23);
\end{tikzpicture}}
    \vspace{-10pt}
    \caption{Admission-aware two-phase readiness for newly registered adapters. Immediate first-touch exposes cold loading to users and disrupts old warm traffic. MinT instead registers the adapter, loads it through the same admitted prewarm path, marks it ready only after activation, and then admits user requests. The zero load p95 applies to the ready path; the cost is the measured background prewarm span before exposure.}
    \vspace{-10pt}
    \label{fig:e4_1m_serving_controls}
\end{figure}

The serving experiments use Qwen3-30B rank-1 MoE LoRA adapters on one 4-GPU tensor-parallel serving actor, with prompt length 1024 and maximum output length 64. The probes cover the path from request name to generation: a materialized 1M packed catalog, repeated-hotset and unique-adapter traffic, warm and cold latency, open-loop Poisson SLO capacity, the tested same-batch adapter window, CPU-cache growth, cold-load accounting, packed tensor loading, hot catalog reload, and admission-aware two-phase rollout. The main measurements do not imply that $10^6$ adapters are simultaneously resident or active; they show that a million adapter revisions can be built, named, audited, and then selectively served through bounded resident working sets. \Cref{tab:e4_serving_summary} gives the main bounds, while \cref{fig:e4_1m_serving_controls} focuses on the readiness gate that turns the hot-reload finding into a serving contract.

\begin{table}[t]
\centering
\scriptsize
\setlength{\tabcolsep}{3.2pt}
\renewcommand{\arraystretch}{1.10}
\caption{Policy-population serving evidence on Qwen3-30B rank-1 LoRA. Catalog scale, residency scale, warm throughput, and rollout-control scale are separate claims.}
\label{tab:e4_serving_summary}
\fittowidth{%
\begin{tabular}{@{}M{0.18\textwidth}M{0.28\textwidth}M{0.24\textwidth}M{0.24\textwidth}@{}}
\toprule
\apphead Tier & Evidence & Measured bound & Interpretation \\
\midrule
\appgroup{4}{Control-plane addressability}
\appkey{Catalog artifact}
& Built and audited a packed Qwen3-30B rank-1 adapter pool
& 1{,}000{,}000 / 1{,}000{,}000 built; 0 errors; 256 / 256 audit OK across 100 shards
& Million scale is an artifact-backed addressability result, not simultaneous residency. \\
\addlinespace[3pt]

\appgroup{4}{Bounded local residency}
\appkey{CPU adapter cache}
& Repeated-hotset and weak-locality runs on one serving actor
& 369 loaded adapters at a 512-adapter hotset; 550 loaded under 2048-adapter pressure
& One actor can keep hundreds of adapters nearby; weak locality still raises the tail. \\
\addlinespace[3pt]

\appkey{GPU batch window}
& Same-batch distinct-adapter probe
& 64 distinct adapters in the tested decoding batch
& Batch execution is the smallest measured adapter-diversity window. \\
\addlinespace[3pt]

\appgroup{4}{Warm selected-revision serving}
\appkey{Open-loop SLO}
& Poisson traffic over 64 warmed revisions on one TP=4 engine
& 0.5 / 1 / 2 rps keep 100\% TTFT$\le$5s; 4 rps drops to 72.1\% SLO
& The warm serving knee appears between 2 and 4 rps for this workload. \\
\addlinespace[3pt]

\appgroup{4}{Cold-load path and representation}
\appkey{Cold loading}
& Warm/cold $N=64$ comparison and $N=16$ cold-load staircase
& Warm p95 21.35 s; cold-cache p95 199.81 s; $N=16$ staircase 1.375--23.267 s
& Different missing adapters serialize through the engine load path before decoding. \\
\addlinespace[3pt]

\appkey{Packed adapter}
& Serving representation removes small tensor fanout
& 37{,}248 tensors to 672 tensors; 1.05$\times$ byte change; 8.5--8.7$\times$ faster live load
& Packing attacks the cache-miss load slice without changing the policy object. \\
\addlinespace[3pt]

\appgroup{4}{Online rollout control}
\appkey{Two-phase readiness}
& Hot reload combined with admission-controlled prewarm
& Warm p95 24.03 s $\rightarrow$ 9.63 s; warm $>$20 s stalls 10 $\rightarrow$ 0; ready-path load p95 0.00 s after a 409.04 s prewarm span
& New adapters become user-visible after activation, so first user requests do not pay adapter loading. \\
\bottomrule
\end{tabular}%
}
\end{table}

\paragraph{Adapter cache levels.}
MinT treats each exported adapter revision as an addressable policy revision that can move through three cache levels. The durable catalog names policy revisions that can be requested; the CPU cache is local to one serving actor; the GPU batch is the smaller set attached to a decoding step. A cold load moves a policy revision from the durable catalog into the actor before decoding can begin.

Large catalogs split fleet-level routing from engine-local memory. A million accumulated policy revisions is a catalog and routing problem; one engine keeps only a bounded local cache. MinT now backs the catalog side with a built 1M packed pool and an audit that samples all 100 shards. The serving side is measured through selected working sets from that catalog: hundreds of CPU-cached adapters, 64 distinct adapters in one same-batch window, and an open-loop warm-capacity curve that reaches 100\% TTFT$\le$5s SLO through 2 rps before the 4 rps knee. Appendix~\cref{tab:app_memory_loader_accounting} gives the byte and memory accounting behind these cache levels: HBM remains dominated by the resident base model, while CPU memory and adapter object shape determine how many policies can be kept cached near one engine.

\paragraph{Cached adapters and batch diversity.}
Local adapter caches absorb locality before requests touch shared storage. Tenant variants, rollback points, personalization branches, and recent evaluation candidates often recur; broad rollout waves and experiment sweeps have much weaker locality. The service has to support both traffic shapes while keeping CPU cache size separate from same-batch execution.

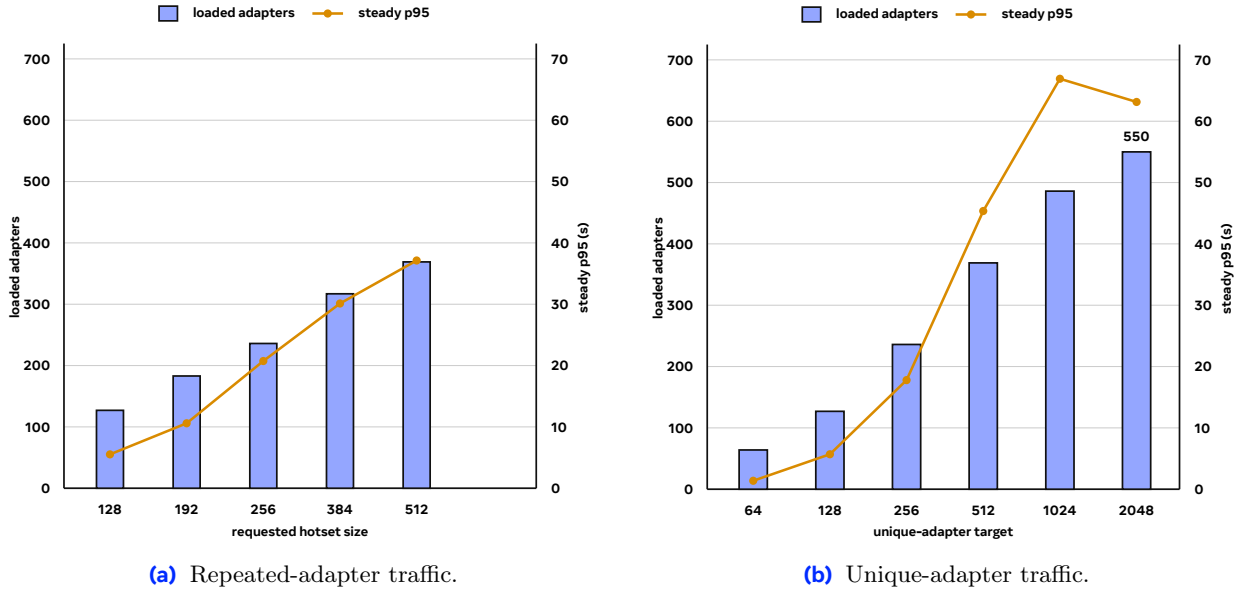
\begin{figure}[t]
    \centering
    \begin{subfigure}[t]{0.48\textwidth}
        \centering
        \resizebox{\linewidth}{!}{\begin{tikzpicture}[x=1cm,y=1cm,>=stealth,font=\sffamily]
  \definecolor{mintauxamber}{HTML}{D98B00}
  \tikzstyle{axis}=[thick, draw=mindlabfg]
  \tikzstyle{bar}=[fill=mindlabblue!42, draw=mindlabfg, thick]
  \tikzstyle{latency}=[draw=mintauxamber, line width=1.2pt]
  \tikzstyle{label}=[font=\scriptsize\sffamily, align=center]

  \foreach \y/\lval/\rval in {0/0/0,1/100/10,2/200/20,3/300/30,4/400/40,5/500/50,6/600/60,7/700/70} {
    \draw[mindlabfg!22] (0,\y) -- (7.7,\y);
    \node[label, anchor=east] at (-0.10,\y) {\lval};
    \node[label, anchor=west] at (7.82,\y) {\rval};
  }
  \draw[axis] (0,0) -- (0,7.25);
  \draw[axis] (0,0) -- (7.7,0);
  \draw[axis] (7.7,0) -- (7.7,7.25);
  \node[label, rotate=90] at (-0.78,3.6) {loaded adapters};
  \node[label, rotate=90] at (8.48,3.6) {steady p95 (s)};

  \foreach \x/\hot/\loaded/\lat in {
    0.75/128/1.27/0.552,
    2.00/192/1.83/1.059,
    3.25/256/2.36/2.074,
    4.50/384/3.17/3.012,
    5.75/512/3.69/3.713
  } {
    \draw[bar] (\x-0.22,0) rectangle (\x+0.22,\loaded);
    \node[label] at (\x,-0.35) {\hot};
  }

  \draw[latency]
    (0.75,0.552) -- (2.00,1.059) -- (3.25,2.074) --
    (4.50,3.012) -- (5.75,3.713);
  \foreach \x/\y in {0.75/0.552,2.00/1.059,3.25/2.074,4.50/3.012,5.75/3.713} {
    \fill[mintauxamber] (\x,\y) circle (1.9pt);
  }

  \node[label] at (3.85,-0.72) {requested hotset size};
  \draw[bar] (1.55,7.62) rectangle (1.85,7.86);
  \node[label, anchor=west] at (1.98,7.74) {loaded adapters};
  \draw[latency] (4.05,7.74) -- (4.55,7.74);
  \fill[mintauxamber] (4.30,7.74) circle (1.9pt);
  \node[label, anchor=west] at (4.68,7.74) {steady p95};
\end{tikzpicture}}
        \caption{Repeated-adapter traffic.}
        \label{fig:e4_hotset_ladder}
    \end{subfigure}\hfill
    \begin{subfigure}[t]{0.48\textwidth}
        \centering
        \resizebox{\linewidth}{!}{\begin{tikzpicture}[x=1cm,y=1cm,>=stealth,font=\sffamily]
  \definecolor{mintauxamber}{HTML}{D98B00}
  \tikzstyle{axis}=[thick, draw=mindlabfg]
  \tikzstyle{bar}=[fill=mindlabblue!42, draw=mindlabfg, thick]
  \tikzstyle{latency}=[draw=mintauxamber, line width=1.2pt]
  \tikzstyle{label}=[font=\scriptsize\sffamily, align=center]

  \foreach \y/\lval/\rval in {0/0/0,1/100/10,2/200/20,3/300/30,4/400/40,5/500/50,6/600/60,7/700/70} {
    \draw[mindlabfg!22] (0,\y) -- (7.7,\y);
    \node[label, anchor=east] at (-0.10,\y) {\lval};
    \node[label, anchor=west] at (7.82,\y) {\rval};
  }
  \draw[axis] (0,0) -- (0,7.25);
  \draw[axis] (0,0) -- (7.7,0);
  \draw[axis] (7.7,0) -- (7.7,7.25);
  \node[label, rotate=90] at (-0.78,3.6) {loaded adapters};
  \node[label, rotate=90] at (8.48,3.6) {steady p95 (s)};

  \foreach \x/\target/\loaded/\lat in {
    0.75/64/0.64/0.138,
    2.00/128/1.27/0.570,
    3.25/256/2.36/1.779,
    4.50/512/3.69/4.536,
    5.75/1024/4.86/6.692,
    7.00/2048/5.50/6.314
  } {
    \draw[bar] (\x-0.23,0) rectangle (\x+0.23,\loaded);
    \node[label] at (\x,-0.35) {\target};
  }

  \draw[latency]
    (0.75,0.138) -- (2.00,0.570) -- (3.25,1.779) --
    (4.50,4.536) -- (5.75,6.692) -- (7.00,6.314);
  \foreach \x/\y in {0.75/0.138,2.00/0.570,3.25/1.779,4.50/4.536,5.75/6.692,7.00/6.314} {
    \fill[mintauxamber] (\x,\y) circle (1.9pt);
  }
  \node[label, anchor=south] at (7.00,5.55) {550};

  \node[label] at (3.85,-0.72) {unique-adapter target};
  \draw[bar] (1.55,7.62) rectangle (1.85,7.86);
  \node[label, anchor=west] at (1.98,7.74) {loaded adapters};
  \draw[latency] (4.05,7.74) -- (4.55,7.74);
  \fill[mintauxamber] (4.30,7.74) circle (1.9pt);
  \node[label, anchor=west] at (4.68,7.74) {steady p95};
\end{tikzpicture}}
        \caption{Unique-adapter traffic.}
        \label{fig:e4_true_unique_frontier}
    \end{subfigure}
    \vspace{-10pt}
    \caption{Cached-adapter scaling on one 4-GPU Qwen3-30B rank-1 serving actor. Bars use the left axis for loaded adapters; lines use the right axis for steady p95 latency. The left panel measures locality absorption under repeated-adapter traffic, reaching 369 loaded adapters at a 512-adapter working set. The right panel measures cache growth under weak locality, reaching 550 loaded adapters at a 2048-adapter unique target while p95 rises to 63.14 s.}
    \vspace{-10pt}
    \label{fig:e4_cache_ladders}
\end{figure}

\Cref{fig:e4_cache_ladders} measures these two regimes. Repeated-adapter traffic reaches 369 loaded adapters at a 512-adapter hotset with p95 37.13 s and no errors. Weak-locality traffic reaches 550 loaded adapters at a 2048-adapter unique target with p95 63.14 s and no errors, so it is cache-growth evidence under pressure rather than a clean warm-cache endpoint. Up to target 512 the two panels report essentially the same loaded count (127, 236, and 369), so locality matters only once the request stream exceeds the working set; beyond that, weak-locality p95 saturates rather than diverges, peaking at 66.92 s at target 1024 and settling at 63.14 s at target 2048 as the CPU cache plateaus near 550. Both loaded-adapter counts are larger than the tested same-batch window of 64 distinct adapters: one actor can cache more adapters than it can execute together in one GPU batch. Routing should preserve warm reuse, while batching still has to respect the smaller same-batch adapter limit. Appendix~\cref{tab:app_cache_ladders} gives the ordered ladder data, and Appendix~\cref{tab:app_business_traffic} shows stress probes where long outputs, weak locality, and high concurrency make simple warm-cache assumptions break down.

The open-loop Poisson sweep measures the selected-revision serving contract under a warmed 1M catalog. With 64 warmed adapters, the TP=4 actor keeps 100\% of requests under the TTFT$\le$5s SLO at 0.5, 1, and 2 target rps; achieved goodput is 0.489, 0.994, and 2.039 rps, respectively. At 4 target rps the actor still completes all 688 submitted requests, but TTFT p95 rises to 5.795 s and SLO attainment falls to 72.1\%, locating the knee between 2 and 4 rps for this workload. The 6 rps overload probe remains useful as a saturation point, not as a capacity claim. Appendix~\cref{tab:app_openloop_rps} reports the full sweep.

\paragraph{Cold loading as service work.}
Cold misses expose the cache-miss work inside a policy population. New adapters, rollback points, long-tail personalization branches, and evaluation snapshots keep producing cache misses. A cold miss fetches adapter tensors, builds loader objects, registers the adapter with the serving engine, and only then starts generation. MinT therefore treats cold loading as scheduled service work, with deduplication for repeated cache-miss requests to the same policy and bounded backpressure when too many distinct missing policies arrive together.

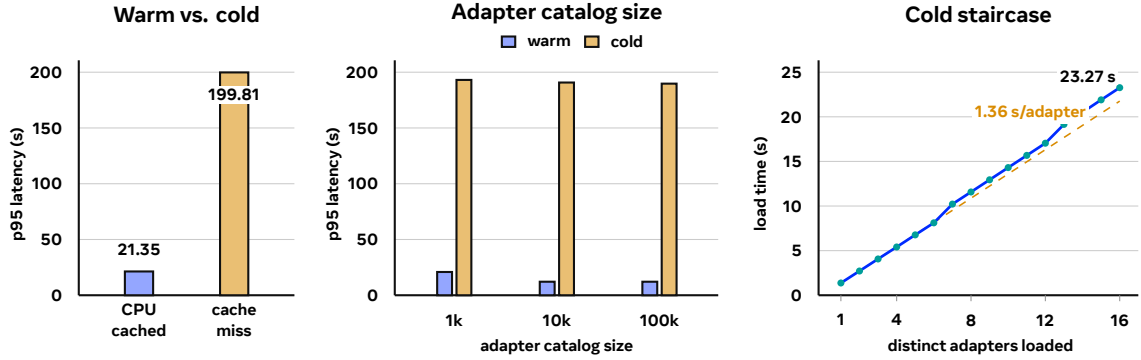
\begin{figure}[t]
    \centering
    \resizebox{0.92\textwidth}{!}{\begin{tikzpicture}[x=1cm,y=1cm,>=stealth,font=\sffamily]
  \definecolor{mintauxteal}{HTML}{009E9A}
  \definecolor{mintauxamber}{HTML}{D98B00}
  \tikzstyle{axis}=[thick, draw=mindlabfg]
  \tikzstyle{warmbar}=[fill=mindlabblue!42, draw=mindlabfg, thick]
  \tikzstyle{coldbar}=[fill=mintauxamber!55, draw=mindlabfg, thick]
  \tikzstyle{observed}=[draw=mindlabblue, line width=1.05pt]
  \tikzstyle{refline}=[draw=mintauxamber, dashed, semithick]
  \tikzstyle{label}=[font=\scriptsize\sffamily, align=center]

  \path[use as bounding box] (-1.05,-0.95) rectangle (14.35,4.12);

  \begin{scope}[shift={(0,0)}]
    \foreach \y/\lab in {0/0,0.75/50,1.50/100,2.25/150,3.00/200} {
      \draw[mindlabfg!22] (0,\y) -- (2.90,\y);
      \node[label, anchor=east] at (-0.10,\y) {\lab};
    }
    \draw[axis] (0,0) -- (2.90,0);
    \draw[axis] (0,0) -- (0,3.16);
    \node[label, rotate=90] at (-0.82,1.50) {p95 latency (s)};
    \node[label] at (1.45,3.78) {\textbf{Warm vs. cold}};

    \draw[warmbar] (0.62,0) rectangle (1.00,0.320);
    \draw[coldbar] (1.90,0) rectangle (2.28,2.997);
    \node[label] at (0.81,-0.33) {CPU\\cached};
    \node[label] at (2.09,-0.33) {cache\\miss};
    \node[label, anchor=south] at (0.81,0.42) {21.35};
    \node[label, fill=white, inner sep=1pt] at (2.09,2.70) {199.81};
  \end{scope}

  \begin{scope}[shift={(4.25,0)}]
    \foreach \y/\lab in {0/0,0.75/50,1.50/100,2.25/150,3.00/200} {
      \draw[mindlabfg!22] (0,\y) -- (4.35,\y);
      \node[label, anchor=east] at (-0.10,\y) {\lab};
    }
    \draw[axis] (0,0) -- (4.35,0);
    \draw[axis] (0,0) -- (0,3.16);
    \node[label, rotate=90] at (-0.82,1.50) {p95 latency (s)};
    \node[label] at (2.18,3.78) {\textbf{Adapter catalog size}};
    \draw[warmbar] (1.44,3.34) rectangle (1.62,3.48);
    \node[label, anchor=west] at (1.66,3.41) {warm};
    \draw[coldbar] (2.56,3.34) rectangle (2.74,3.48);
    \node[label, anchor=west] at (2.78,3.41) {cold};

    \foreach \x/\pool/\warmheight/\coldheight/\warmval/\coldval in {
      0.80/1k/0.313/2.896/20.9/193.0,
      2.18/10k/0.182/2.861/12.2/190.7,
      3.56/100k/0.182/2.846/12.1/189.7
    } {
      \draw[warmbar] (\x-0.23,0) rectangle (\x-0.03,\warmheight);
      \draw[coldbar] (\x+0.03,0) rectangle (\x+0.23,\coldheight);
      \node[label] at (\x,-0.33) {\pool};
    }
    \node[label] at (2.18,-0.70) {adapter catalog size};
  \end{scope}

  \begin{scope}[shift={(9.95,0)}]
    \foreach \y/\lab in {0/0,0.60/5,1.20/10,1.80/15,2.40/20,3.00/25} {
      \draw[mindlabfg!22] (0,\y) -- (4.30,\y);
      \node[label, anchor=east] at (-0.10,\y) {\lab};
    }
    \draw[axis] (0,0) -- (4.30,0);
    \draw[axis] (0,0) -- (0,3.16);
    \node[label, rotate=90] at (-0.78,1.50) {load time (s)};
    \node[label] at (2.15,3.78) {\textbf{Cold staircase}};

    \foreach \x/\lab in {0.30/1,1.05/4,2.05/8,3.05/12,4.05/16} {
      \draw[mindlabfg!45] (\x,0) -- (\x,-0.08);
      \node[label] at (\x,-0.33) {\lab};
    }
    \node[label] at (2.18,-0.70) {distinct adapters loaded};

    \draw[refline] (0.30,0.163) -- (4.05,2.611);
    \draw[observed]
      (0.30,0.165) -- (0.55,0.326) -- (0.80,0.488) -- (1.05,0.649)
      -- (1.30,0.812) -- (1.55,0.974) -- (1.80,1.226) -- (2.05,1.389)
      -- (2.30,1.553) -- (2.55,1.717) -- (2.80,1.881) -- (3.05,2.045)
      -- (3.30,2.298) -- (3.55,2.461) -- (3.80,2.628) -- (4.05,2.792);
    \foreach \x/\y in {
      0.30/0.165,0.55/0.326,0.80/0.488,1.05/0.649,
      1.30/0.812,1.55/0.974,1.80/1.226,2.05/1.389,
      2.30/1.553,2.55/1.717,2.80/1.881,3.05/2.045,
      3.30/2.298,3.55/2.461,3.80/2.628,4.05/2.792
    } {
      \fill[mintauxteal] (\x,\y) circle (1.35pt);
    }
    \node[label, text=mintauxamber, fill=white, inner sep=1pt] at (2.85,2.45) {1.36 s/adapter};
    \node[label, fill=white, inner sep=1pt] at (3.62,2.95) {23.27 s};
  \end{scope}
\end{tikzpicture}}
    \vspace{-10pt}
    \caption{Warm routing and bounded cold-load cost. Left: CPU-cached adapters take the warm latency path, while cache-miss policies take the cold path. Middle: the earlier 1k--100k sweep keeps warm measurements in the warm p95-latency regime and cold measurements in the cold p95-latency regime as namespace size grows. Right: 16 different cache-miss policies form a serialized load staircase from 1.375 s to 23.267 s, about 1.35--1.40 s per policy.}
    \vspace{-10pt}
    \label{fig:e4_latency_catalog}
\end{figure}

\Cref{fig:e4_latency_catalog} shows why cold loading is a separate capacity dimension. CPU-cached requests take the warm regime. Cache-miss requests take the cold regime. The earlier 1k--100k catalog sweep keeps the same warm/cold latency modes as namespace size grows; Appendix~\cref{tab:app_path_pool_sweep} reports that sweep as supporting evidence for the residency boundary, while Appendix~\cref{tab:app_1m_catalog_audit} now carries the million-scale artifact claim. The right panel isolates the controlled cold-miss component: 16 different cache-miss policies form a load staircase from 1.375 s to 23.267 s, about 1.35--1.40 s per policy. Concurrent cache-miss requests for the same missing policy can share one load, while different missing policies remain separate load jobs. Appendix~\cref{tab:app_cold_load_control} decomposes this path into API queueing, shared loads, unique-policy loading, and retryable load rejection.

\paragraph{Hot reload and readiness.}
Online catalog updates split into registration and activation. In the hot-reload probe, MinT starts with 32 registered and warmed adapters, sends old warm traffic at 0.5 rps, registers 16 new adapters while serving continues, and then immediately first-touches the new adapters. Registration itself is fast: the 16-adapter hot registration takes 1.66 s total with p95 0.136 s. The dangerous part is activation. Existing warm traffic remains error-free, but post-reload TTFT p95 rises from 0.549 s to 20.840 s while new adapters pay first-touch TTFT p95 14.446 s and load p95 22.205 s. This experiment changes the design target: the serving layer has to optimize activation and exposure, not merely catalog registration.

\begin{table}[t]
\centering
\scriptsize
\setlength{\tabcolsep}{3.2pt}
\renewcommand{\arraystretch}{1.10}
\caption{Mixed warm/cold hot-reload control on a 1M packed catalog. Admission protects old warm tenants by moving cold activation into a scheduled path; two-phase readiness then exposes new adapters only after activation, so zero load time applies to ready-path user requests rather than registration time.}
\label{tab:e4_rollout_control}
\fittowidth{%
\begin{tabular}{@{}M{0.22\textwidth}M{0.23\textwidth}M{0.29\textwidth}M{0.20\textwidth}@{}}
\toprule
\apphead Policy & Existing warm traffic & New-adapter path & Interpretation \\
\midrule
\appgroup{4}{Expose before readiness}
\appkey{Admission off}
& post TTFT p95 24.03 s; $>$20 s stalls: 10
& e2e p95 59.18 s; user TTFT p95 22.19 s; load p95 47.37 s
& Fast exposure, but cold first-touch disrupts warm tenants. \\
\addlinespace[3pt]
\appkey{Admission on}
& post TTFT p95 9.71 s; $>$20 s stalls: 0
& e2e p95 314.79 s; user TTFT p95 10.68 s; load p95 294.96 s
& Admission protects warm tenants, but new users wait behind activation. \\
\addlinespace[3pt]
\appgroup{4}{Expose after readiness}
\appkey{Two-phase readiness}
& post TTFT p95 9.63 s; $>$20 s stalls: 0
& ready-path TTFT p95 4.60 s; load p95 0.00 s; prewarm span 409.04 s
& First user requests arrive after activation, so they do not load adapters. \\
\bottomrule
\end{tabular}%
}
\end{table}

\Cref{tab:e4_rollout_control} reports the mixed warm/cold admission experiment and the resulting optimization. Admission control does not make cold loading faster; it protects existing tenants by bounding how much cold activation enters the engine at once. Without admission, old warm traffic sees post-reload TTFT p95 24.03 s and 10 stalls above 20 s. With admission, old warm TTFT p95 falls to 9.71 s and the $>$20 s stalls disappear, but new cold first-touch load p95 rises to 294.96 s because those requests wait for activation. The production contract should therefore not expose a newly registered adapter immediately. MinT changes the visibility boundary: register the adapter, prewarm it through the same cold-load admission path, mark it ready only after activation, and then admit user traffic. This keeps the old warm p95 at 9.63 s with no $>$20 s stalls while the first user-visible request to the new adapter has load p95 0.00 s and TTFT p95 4.60 s. The 0.00 s number is a ready-path result, not a zero-wait deployment result: in this probe, 32 adapters require a 409.04 s background prewarm span before they are marked ready.

\paragraph{Cold-load representation.}
The load staircase also explains why the adapter file format matters. A rank-1 MoE LoRA adapter is moderate in bytes, while the measured adapter file is fragmented into 37{,}248 tensor objects, mostly tiny expert tensors. In these probes, local-disk staging shortened the read path; it left tensor fanout, Python object creation, and loader-side registration work unchanged. MinT therefore packs MoE expert tensors into a serving representation with nearly unchanged declared bytes, reducing object fanout before the engine loads the policy.

\begin{table}[t]
\centering
\scriptsize
\setlength{\tabcolsep}{4.0pt}
\renewcommand{\arraystretch}{1.08}
\caption{Packed MoE LoRA loading reduces cold-load overhead by removing tensor fanout. The byte-size change is small; the speedup comes from replacing many tiny tensor objects with a compact serving representation. The final three rows report the median duration of one live load call measured at each cache-miss burst size.}
\label{tab:e4_packed_loader}
\fittowidth{%
\begin{tabular}{@{}M{0.32\textwidth}M{0.16\textwidth}M{0.16\textwidth}M{0.24\textwidth}@{}}
\toprule
\apphead Metric & Original & Packed & Effect \\
\midrule
\appgroup{4}{Adapter-file shape}
\appkey{File size} & 110.75 MB & 105.58 MB & 1.05$\times$ smaller \\
\appkey{Tensor objects} & 37{,}248 & 672 & 55.4$\times$ fewer \\
\appgroup{4}{Cold-load slice}
\appkey{Read tensors} & 0.3669 s & 0.0067 s & 54.8$\times$ faster \\
\appkey{Build loader objects} & 0.7540 s & 0.0256 s & 29.5$\times$ faster \\
\appgroup{4}{Live engine loading}
\appkey{$N=4$ live load} & 1.363 s & 0.156 s & 8.7$\times$ faster \\
\appkey{$N=8$ live load} & 1.361 s & 0.159 s & 8.6$\times$ faster \\
\appkey{$N=16$ live load} & 1.388 s & 0.164 s & 8.5$\times$ faster \\
\bottomrule
\end{tabular}%
}
\end{table}

\Cref{tab:e4_packed_loader} shows the effect. Packing changes file size only from 110.75 MB to 105.58 MB and reduces tensor objects from 37{,}248 to 672. The direct loader slice improves by 29.5--54.8$\times$, and live engine loading for $N=4$, $N=8$, and $N=16$ improves by 8.5--8.7$\times$, with packed live-load medians under 0.2 seconds. This number refers to the live engine-load slice after packing; end-to-end cold latency still includes routing, queueing, fetch, and generation. The measured speedup comes from object layout: the serving representation removes the small-object storm on cache misses while the declared tensor bytes stay nearly unchanged.

\FloatBarrier
Together, these measurements locate the serving work for policy populations. The 1M catalog is the durable addressability scale; CPU-cache size, tested same-batch diversity, warm open-loop SLO capacity, unique cold-load cost, and readiness state are online execution scales. Routing and prewarming turn recurring traffic into CPU-cache hits. Batch construction respects the smaller same-batch adapter limit. Cache misses enter an explicit cold-load stage before sampling latency begins. Packing reduces the load staircase itself, while admission-aware two-phase rollout decides when a newly registered adapter becomes user-visible. Neither optimization changes the same-batch adapter window; together they reduce activation cost and isolate the remaining waiting time from warm traffic. Appendix~\cref{app:serving} contains the full ladders, accounting tables, stress probes, audit data, native-vLLM caveat, and rollout-control probes behind this contract.
\FloatBarrier

\section{Related Work}
\label{sec:related}

\paragraph{Post-training workload.}
MinT targets a workload where post-training leaves many policy variants over a smaller set of base models. Yao argues that future progress shifts more weight toward problem definition, evaluation, and agent-environment interaction~\citep{yao_second_half_2025}. Silver and Sutton frame future agents as learning from streams of experience in addition to static human data~\citep{silver_sutton_experience_2025}. Recent frontier-model reports expose the same systems pressure through reasoning, coding, tool use, executable environments, asynchronous agent RL, and long-horizon evaluation~\citep{deepseek_v4_release_2026,deepseekv4_2026,glm5_2026,kimi_k25_2026,minimax_m27_2026,qwen35_2026,openai_gpt55_2026,anthropic_opus45_2025,anthropic_agent_autonomy_2026}. These workloads produce task variants, evaluation candidates, product branches, tenant adapters, rollback points, and personal adapters that share base models while retaining separate histories.

\paragraph{Service interfaces.}
Remote post-training interfaces make training loops programmable. Tinker exposes low-level post-training primitives through a remote service~\citep{tinker2025,tinker_cookbook}. SkyRL tx implements a Tinker-style backend and documents model and benchmark coverage across Qwen3 dense and MoE models, Llama 3, DeepSeek V3, GSM8K, and DAPO/AIME recipes~\citep{skyrl_tx}. OpenTinker frames a broader RL-as-a-Service stack around agents, environments, protocols, scheduling, training, and inference~\citep{opentinker2026}. Tinker specifies how users program remote post-training; MinT keeps a compatible programming surface and defines the LoRA-specific service path underneath it: policy record, training checkpoint, rollout record, exported adapter revision, and adapter cache state.

\paragraph{RL execution systems.}
HybridFlow/verl, AReaL, OpenRLHF, Relax, ROLL, StreamRL, AsyncFlow, Laminar, and NeMo-Aligner study rollout scheduling, colocated and disaggregated execution, asynchronous optimization, GPU utilization, failure isolation, and end-task quality~\citep{hybridflow2025,areal2025,openrlhf2024,relax2026,roll2025,streamrl2025,asyncflow2025,laminar2025,nemoaligner2024}. Their central systems objects are actors, rollout replicas, parameter services, queues, and placement groups. MinT uses the same execution concerns and adds LoRA-specific service state: adapter revisions, optimizer state, rollout records, MoE route records, DSA correction metadata, exported adapters, and adapter cache state stay tied to the policy that produced them.

\paragraph{Training-serving consistency.}
Modern RL pipelines often generate rollouts through an inference engine and score them through a training backend. Yao et al. study token-probability gaps in hybrid vLLM/FSDP pipelines and propose truncated importance sampling~\citep{rollout_training_mismatch2025}. Jet-RL shows that mixed BF16 training and FP8 rollout can destabilize on-policy RL under long generations, then uses a unified precision flow to reduce numerical mismatch~\citep{jetrl2026}. R3 studies MoE router disagreement between training and inference and reuses inference-time expert-route records during training~\citep{r3_moe_router2025}. MinT addresses the same problem by storing the adapter revision, rollout record, MoE route or DSA correction metadata, and served evaluation path that name which behavior generated and scored a token.

\paragraph{Parameter-efficient tuning and multi-LoRA training.}
LoRA freezes the base model and trains low-rank matrices attached to selected layers~\citep{lora2022}. AdaLoRA allocates rank budget across matrices according to importance~\citep{adalora2023}. QLoRA combines frozen quantized bases with LoRA updates for memory-efficient fine-tuning~\citep{qlora2023}. LoRA Without Regret argues that LoRA can reach strong post-training quality~\citep{lora_without_regret_2025}, supporting the premise that LoRA can be more than a memory-saving approximation. MinT turns that premise into managed infrastructure for RL service operation: each update restores one adapter and optimizer state over a resident base, each export produces a serving revision, and each rollout or serving request selects that revision through cache and residency state.

\paragraph{Multi-LoRA serving.}
Multi-LoRA serving systems optimize inference after adapters already exist. Punica, S-LoRA, dLoRA, dynamic operator optimization, and vLLM improve batching, memory management, scheduling, and kernels for many adapters over a shared base~\citep{punica2024,slora2023,dlora2024,zhou2025dynamic,vllm2023}. Compress then Serve reduces adapter storage and serving overhead through individual and joint LoRA compression~\citep{compress_then_serve2024}. FastLibra manages LoRA and KV cache dependencies in a unified HBM cache~\citep{fastlibra2025}. LoRAServe handles rank heterogeneity, placement, and routing across distributed LoRA serving clusters~\citep{loraserve2025}. MinT connects the serving catalog to the training lifecycle: an addressable adapter is an exported revision of a trained policy, and a cache miss reloads that revision from the policy's exported adapter file.

\paragraph{Large-model infrastructure.}
Ray provides the distributed execution framework used by many AI systems~\citep{ray2018}. Megatron-LM, Efficient Megatron-LM, ZeRO, and MoE Parallel Folding provide the model-parallel, memory-partitioning, and MoE-parallel techniques that make large-model and sparse-MoE training practical~\citep{megatronlm2019,efficient_megatronlm2021,zero2019,moe_parallel_folding2026}. DeepSeek-V4 describes large-model infrastructure through concrete state units such as hybrid KV cache entries, state caches, on-disk cache storage, rollout WALs, and teacher-state scheduling~\citep{deepseekv4_2026}. MinT applies the same resource-accounting style to LoRA RL: base deployments stay resident, adapter revisions move across subsystems, and adapter cache state controls which exported policies are hot.

\section{Conclusion}

Post-training now couples rollout, update, evaluation, serving, scheduling, and data movement around many policy variants over a small number of expensive base-model deployments. Full-checkpoint workflows make every task branch, benchmark candidate, product version, tenant variant, rollback point, or personal policy look like another complete model deployment. This abstraction does not scale when the target workload is a large and continuously changing population of trained behaviors over shared frontier bases.

MinT makes exported LoRA adapter revisions the managed unit for this setting. Base-model deployments remain resident, while adapter revisions carry trained behavior through rollout, update, export, evaluation, serving, and rollback. The adapter revision is the behavior-carrying payload; the policy record is the service state that makes that payload reproducible, schedulable, and durable across worker changes, training checkpoints, rollout records, and serving-cache movement.

The measurements validate the same adapter-revision path along three scaling axes. \textit{Scale Up} supports LoRA RL on dense and MoE architectures with model-parallel training and serving paths exercised beyond 1T total parameters. \textit{Scale Down} removes full-checkpoint materialization from the training-serving handoff: adapter-only handoff reduces the measured handoff step by $18.3\times$ on a 4B dense model and by $2.85\times$ on a 30B MoE model, while concurrent multi-policy GRPO shortens wall time by $1.77\times$ and $1.45\times$ under the same resident-base allocation. \textit{Scale Out} separates durable policy addressability from CPU/GPU hot working sets with an artifact-backed 1M catalog: MinT builds 1{,}000{,}000 packed adapter revisions with zero errors, audits sampled adapters across all 100 shards, and serves selected revisions through bounded local working sets. Warm open-loop serving keeps 100\% TTFT$\le$5s through 2 rps for the measured 64-adapter workload, while hot-reload experiments show that the dominant online risk is cold activation interference rather than catalog registration. Packed MoE LoRA tensors reduce live-load fanout by $8.5$--$8.7\times$, and admission-aware two-phase rollout protects old warm tenants while moving adapter loading before readiness, so ready-path user first requests avoid load cost at the price of explicit prewarm delay.

Together, these results show that multi-tenant LoRA training services can be made practical without turning every trained policy into a full-model server. MinT lets policy count grow through adapter revisions, policy records, cache tiers, and controlled readiness state, while selected revisions train and serve over bounded resident working sets. This makes LoRA a service-level unit for large-scale post-training today and a practical path toward larger populations of organizational and personal policies over shared 1T-class base models.

\bibliographystyle{plainnat}
\bibliography{paper}

\newpage
\begin{appendices}
\section{Author List}
\label{app:author_list}

Names are listed alphabetically.

\paragraph{Core Contributors.}
\begin{sloppypar}
Andrew Chen, Cleon Cheng, Steven Chiang, Nolan Ho, Andrew Lei, Lucian Li, Kieran Liu, Irvine Lu, Pony Ma, Rio Yang, Di Zhang, Adrian Zhou.
\end{sloppypar}

\paragraph{Team.}
\begin{sloppypar}
Song Cao, Vic Cao, Kaijie Chen, Bunny Fan, Hera Feng, Huan Feng, Arthur Fu, Jun Gao, Hongquan Gu, Aaron Guan, Mutian Hong, Hailee Hou, Peixuan Hua, Charles Huang, Miles Jiang, Nora Jiang, Yuyi Jiang, Autumn Jin, Fancy Kong, Kyrie Lei, Alexy Li, Dawn Li, Ray Li, Theo Li, Jiayi Lin, Domini Liu, Heshan Liu, Kairus Liu, Logan Liu, Maeve Luo, Runism Lv, Pony Ma, Verity Niu, Anson Qiu, Vincent Wang, Maxwell Yao, Regis Ye, Wenlin Ye, Yanying Ye, Josh Ying, Danney Zeng, Salmon Zhan, Anya Zhang, Ruijia Zhang, Sueky Zhang, Ya Zhang, Wei Zhao, Ada Zhou, Sizer Zhou, Xinyue Zhu, Murphy Zhuang.
\end{sloppypar}

\section{Additional Serving Measurements}
\label{app:serving}

This appendix expands the serving-side measurements used in \cref{sec:e4_serving}. The tables use the same serving quantities as the main text: catalog size, CPU-cached adapters, adapters in one GPU batch, and the cold-load path that loads an adapter when a request selects an entry outside the actor's local cache. The order separates exploratory probes that rule out simple explanations, measurements that identify the serving limit, and follow-up probes that reduce or isolate that limit.

Several rows intentionally leave the clean warm path. They create weak locality, cold churn, long online generations, or high concurrency so that the failure mode is visible rather than hidden inside average latency. We keep them in the appendix because they identify the control points that a MinT deployment must own: routing for warm reuse, prewarming for predictable waves, bounded cold loading for cache-miss traffic, and lower-fanout adapter representation for unavoidable misses. The shaded bands in the tables separate cache levels and probe families, so a reader can scan from the MinT service path to the measured limit without treating every row as the same kind of claim.

\paragraph{One-million-adapter catalog artifact.}
The main text uses the 1M catalog as an artifact-backed addressability claim. MinT generated a packed Qwen3-30B-A3B rank-1 catalog under \texttt{/tos-mindverse/zhouch\_pool\_30b\_r1\_bf16\_v1/pool\_30b\_r1\_bf16\_packed\_v2}. The build produced 1{,}000{,}000 adapters with zero build errors across 100 shards. A fresh audit sampled 256 adapters across all 100 shards, opened the packed safetensors files, and read tensors from the sampled adapters.

\begin{table}[H]
\centering
\scriptsize
\setlength{\tabcolsep}{3.2pt}
\renewcommand{\arraystretch}{1.08}
\caption{One-million packed LoRA catalog build and audit. The audit result is the evidence that the 1M catalog is a built and readable artifact, not a namespace-only sizing model.}
\label{tab:app_1m_catalog_audit}
\fittowidth{%
\begin{tabular}{@{}M{0.24\textwidth}M{0.27\textwidth}M{0.38\textwidth}@{}}
\toprule
\apphead Item & Measurement & Evidence / interpretation \\
\midrule
Adapters built & 1{,}000{,}000 / 1{,}000{,}000 & Packed adapter revisions materialized on TOS \\
Build errors & 0 & Build completed without failed adapter rows \\
Fleet wall time & 138{,}521.2 s & 38.5 h end-to-end build wall time \\
Aggregate write throughput & 762.2 MB/s & Fleet-level write throughput during build \\
Shard layout & 100 shards, 10{,}000 adapters/shard & Durable catalog layout for addressability \\
Audit sample & 256 / 256 OK, 0 errors & Samples span all 100 shards \\
Audit wall time & 144.81 s & Lightweight read audit after build \\
Audit schema sample & \makecell{packed v0 index\\288 groups; 384 copied\\672 safetensors keys} & Confirms packed serving representation \\
Per-adapter audit elapsed & p50 8.54 s; p95 10.96 s; max 12.21 s & Includes safetensors open and sampled tensor reads \\
\bottomrule
\end{tabular}%
}
\end{table}

\paragraph{Open-loop warm capacity on the 1M catalog.}
The open-loop sweep uses 64 warmed adapters selected from the 1M packed catalog. Arrivals follow a Poisson process for 180 s, with prompt length 1024, maximum output length 64, and TTFT$\le$5 s as the SLO. The sweep is a selected-revision capacity measurement, not a claim that all 1M adapters are resident.

\begin{table}[H]
\centering
\scriptsize
\setlength{\tabcolsep}{3.2pt}
\renewcommand{\arraystretch}{1.08}
\caption{Open-loop Poisson capacity sweep for warmed selected revisions from the 1M packed catalog.}
\label{tab:app_openloop_rps}
\fittowidth{%
\begin{tabular}{@{}ccccccc@{}}
\toprule
\apphead Target RPS & Achieved RPS & OK / submitted & TTFT p50 & TTFT p95 & SLO attain & SLO goodput \\
\midrule
0.5 & 0.489 & 88 / 88 & 0.526 s & 0.627 s & 100.0\% & 0.489 rps \\
1.0 & 0.994 & 179 / 179 & 0.511 s & 0.617 s & 100.0\% & 0.994 rps \\
2.0 & 2.039 & 367 / 367 & 0.522 s & 0.653 s & 100.0\% & 2.039 rps \\
4.0 & 3.822 & 688 / 688 & 3.322 s & 5.795 s & 72.1\% & 2.756 rps \\
6.0 & 5.750 & 1035 / 1035 & 2.871 s & 5.465 s & 68.1\% & 3.917 rps \\
\bottomrule
\end{tabular}%
}
\end{table}

\paragraph{Hot reload, admission, and readiness probes.}
\Cref{tab:app_hot_reload_rollout} gives the detailed rows behind the main rollout-control result. Registration is fast in both hot-reload probes; cold activation is the operation that interferes with warm traffic. Admission serializes cold work to protect existing tenants. Two-phase readiness moves that waiting time into a rollout/prewarm phase before user traffic can select the new adapter; a zero user-visible load p95 is therefore a ready-path measurement, not a zero-time deployment claim.

\begin{table}[H]
\centering
\scriptsize
\setlength{\tabcolsep}{2.8pt}
\renewcommand{\arraystretch}{1.06}
\caption{Hot reload and mixed warm/cold rollout-control measurements. Existing traffic is old warmed-adapter traffic; added traffic refers to newly registered adapters.}
\label{tab:app_hot_reload_rollout}
\fittowidth{%
\begin{tabular}{@{}M{0.20\textwidth}cccccM{0.20\textwidth}@{}}
\toprule
\apphead Probe & Existing OK & Existing post TTFT p95 & Existing $>$20 s & Added e2e p95 & Added user TTFT p95 / load p95 & Readout \\
\midrule
Hot reload, immediate first-touch & 40 / 40 & 20.840 s & -- & 36.785 s & 14.446 s / 22.205 s & Registration 16 adapters takes 1.66 s; activation causes the tail. \\
Admission off & 148 / 148 & 24.028 s & 10 & 59.181 s & 22.195 s / 47.370 s & Cold first-touch interferes with warm tenants. \\
Admission on & 148 / 148 & 9.706 s & 0 & 314.794 s & 10.677 s / 294.961 s & Cold work waits; old warm traffic is protected. \\
Admission on + two-phase prewarm & 148 / 148 & 9.630 s & 0 & 21.430 s & 4.600 s / 0.000 s & Users see only ready adapters after a 409.04 s prewarm span; load cost moves before exposure. \\
\bottomrule
\end{tabular}%
}
\end{table}

\paragraph{Native vLLM baseline and caveat.}
The native vLLM experiment has two different meanings. Directly loading the MinT 1M packed adapter layout into native vLLM fails because the packed directory layout is not a standard PEFT tensor layout, so this row is a compatibility boundary rather than a latency comparison. A separate standard-PEFT 1k baseline shows that native vLLM performs well on small static PEFT adapter sets; it is not an apples-to-apples measurement of MinT's 1M packed catalog lifecycle.

\begin{table}[H]
\centering
\scriptsize
\setlength{\tabcolsep}{3.0pt}
\renewcommand{\arraystretch}{1.08}
\caption{Native vLLM caveat. Standard PEFT baseline rows are useful context, while the 1M packed row is a format-compatibility failure rather than a performance result.}
\label{tab:app_native_vllm_caveat}
\fittowidth{%
\begin{tabular}{@{}M{0.22\textwidth}cccccM{0.23\textwidth}@{}}
\toprule
\apphead Setup & Target RPS & Achieved RPS & OK / submitted & TTFT p50 & TTFT p95 & Interpretation \\
\midrule
MinT 1M packed layout & -- & -- & failed startup & -- & -- & Native vLLM reports that the adapter directory does not contain standard tensors. \\
Standard PEFT 1k baseline & 0.5 & 0.517 & 62 / 62 & 0.474 s & 0.561 s & Static small-pool PEFT baseline. \\
Standard PEFT 1k baseline & 1.0 & 1.033 & 124 / 124 & 0.484 s & 0.571 s & Static small-pool PEFT baseline. \\
Standard PEFT 1k baseline & 2.0 & 2.017 & 242 / 242 & 0.530 s & 0.653 s & Static small-pool PEFT baseline. \\
Standard PEFT 1k baseline & 4.0 & 3.817 & 458 / 458 & 0.535 s & 0.762 s & Static small-pool PEFT baseline. \\
\bottomrule
\end{tabular}%
}
\end{table}

\paragraph{Adaptive activation scheduler negative result.}
After the two-phase rollout result, we tested a more aggressive SLO-aware activation scheduler that changes prewarm concurrency from recent warm-traffic TTFT. The result is intentionally not used as a positive claim. It shortens rollout wall time, but the aggressive variants hurt warm tails, identifying executor/add-LoRA activation interference as the next bottleneck rather than something solved by higher cold-load concurrency. This negative result is why the main paper reports the conservative two-phase gate instead of treating higher prewarm parallelism as an optimization.

\begin{table}[H]
\centering
\scriptsize
\setlength{\tabcolsep}{3.0pt}
\renewcommand{\arraystretch}{1.06}
\caption{Adaptive activation scheduler probes. The conservative fixed policy remains the paper-safe rollout control.}
\label{tab:app_adaptive_scheduler}
\fittowidth{%
\begin{tabular}{@{}M{0.22\textwidth}cccccM{0.18\textwidth}@{}}
\toprule
\apphead Policy & Prewarm span & Existing post TTFT p95 & Existing $>$20 s & User added TTFT p95 & User load p95 & Readout \\
\midrule
Fixed conservative & 409.04 s & 9.63 s & 0 & 4.60 s & 0.00 s & Safe main result. \\
Adaptive cap=1 & 374.33 s & 11.12 s & 0 & 4.77 s & 0.00 s & Small rollout speedup; warm tail slightly worse. \\
Adaptive cap=4 & 358.26 s & 28.63 s & 33 & 4.62 s & 0.00 s & Too aggressive; hurts warm traffic. \\
Traffic-stop drain & 217.69 s & 30.27 s & 23 & 4.57 s & 0.00 s & Fast but unsafe for old traffic. \\
Warm-complete drain & 268.91 s & 22.25 s & 13 & 4.33 s & 0.00 s & Still unsafe for old traffic. \\
\bottomrule
\end{tabular}%
}
\end{table}

\paragraph{Adapter memory and representation.}
\Cref{tab:app_memory_loader_accounting} separates adapter bytes, tensor fanout, CPU cache footprint, and base-model HBM footprint. This accounting explains why the packed loader experiment matters. The measured adapter is moderate in byte size and fragmented into tens of thousands of small tensors, most of them no larger than 4 KB. Cold loading therefore pays object and registration overhead even when the total bytes are small.

\begin{table}[H]
\centering
\scriptsize
\setlength{\tabcolsep}{4pt}
\renewcommand{\arraystretch}{1.08}
\caption{Memory and representation accounting for the measured 30B MoE LoRA adapter files. The rows separate byte size from object fanout and cache pressure.}
\label{tab:app_memory_loader_accounting}
\fittowidth{%
\begin{tabular}{@{}M{0.18\textwidth}M{0.24\textwidth}M{0.27\textwidth}M{0.23\textwidth}@{}}
\toprule
\apphead Tier / object & Item & Measurement & Serving implication \\
\midrule
\appgroup{4}{Adapter-file shape}
Adapter file & Adapter bytes & 110.75 MB file; 105.5 MB declared tensor bytes & Bytes are moderate for a 30B rank-1 adapter. \\
Adapter file & Tensor fanout & 37{,}248 tensors; 37{,}152 no larger than 4 KB & Object fanout dominates cold load work. \\
\appgroup{4}{Resident execution footprint}
CPU warm tier & CPU cache footprint & 586.8 MB/LoRA across 4 TP workers; 146.7 MB/worker & CPU cache is the per-engine cache limit. \\
GPU batch tier & GPU memory & $N=64$ snapshots show about 272 GiB engine HBM footprint & Base deployment dominates HBM pressure. \\
\appgroup{4}{Packed cold-load representation}
Adapter-file format & Packed representation & 37{,}248 $\rightarrow$ 672 tensors; 110.75 $\rightarrow$ 105.58 MB & Packing removes fanout; bytes stay nearly unchanged. \\
\bottomrule
\end{tabular}%
}
\end{table}

\paragraph{Legacy 1k--100k catalog sweep.}
\Cref{tab:app_path_pool_sweep} is retained as the earlier catalog-sweep probe. It varies the adapter catalog from 1k to 100k entries and shows that the warm/cold split persists as catalog names grow. The 1M build/audit table above supersedes this sweep for the million-scale artifact claim; this table remains useful for showing that cache state, not namespace size alone, determines warm versus cold latency.

\begin{table}[H]
\centering
\scriptsize
\setlength{\tabcolsep}{4pt}
\renewcommand{\arraystretch}{1.06}
\caption{Adapter catalog sweep for $N=64$ serving. Warm and cold rows are split so the stable regime gap is visible at each catalog size.}
\label{tab:app_path_pool_sweep}
\fittowidth{%
\begin{tabular}{@{}M{0.13\textwidth}M{0.13\textwidth}ccccM{0.27\textwidth}@{}}
\toprule
\apphead Catalog & Regime & Success & Drain & p95 & Distinct & Readout \\
\midrule
1k & \appkey{warm} & 64/64 & 21.16 s & 20.89 s & 51 & catalog lookup stays in the warm regime \\
1k & \appkey{cold} & 64/64 & 193.23 s & 193.04 s & 56 & cold loading dominates \\
\addlinespace
10k & \appkey{warm} & 64/64 & 12.56 s & 12.16 s & 64 & larger catalog stays in the warm mode \\
10k & \appkey{cold} & 64/64 & 190.99 s & 190.73 s & 56 & cold regime persists \\
\addlinespace
100k & \appkey{warm} & 64/64 & 20.35 s & 12.12 s & 63 & 100k-entry catalog remains addressable \\
100k & \appkey{cold} & 63/64 & 190.08 s & 189.73 s & 56 & tail follows cold loading \\
\bottomrule
\end{tabular}%
}
\end{table}

\paragraph{Cached working sets.}
\Cref{tab:app_cache_ladders} gives the ordered data behind the warm-cache claim in \cref{fig:e4_cache_ladders}. The repeated-hotset rows model adapter locality after routing has found a useful engine placement. The unique-adapter rows remove this locality and measure how many distinct adapters can become cached near one engine before the run stops being a clean warm-path claim. These measurements define the CPU-side tier between the durable adapter catalog and the same-batch adapter window.

\begin{table}[H]
\centering
\scriptsize
\setlength{\tabcolsep}{3.4pt}
\renewcommand{\arraystretch}{1.05}
\caption{Cached-working-set ladder on one 4-GPU Qwen3-30B rank-1 serving actor. Shaded group rows separate routed locality from unique-adapter load pressure.}
\label{tab:app_cache_ladders}
\fittowidth{%
\begin{tabular}{@{}M{0.20\textwidth}cccccM{0.22\textwidth}@{}}
\toprule
\apphead Regime & Target & Loaded & Steady p95 & Errors & Tier & Readout \\
\midrule
\appgroup{7}{Routed locality: repeated hotsets}
Repeated hotsets & 128 & 127 & 5.52 s & 0/3200 & CPU warm & baseline warm tier \\
Repeated hotsets & 192 & 183 & 10.59 s & 0/3200 & CPU warm & locality preserved \\
Repeated hotsets & 256 & 236 & 20.74 s & 0/3200 & CPU warm & p95 rises with hotset \\
Repeated hotsets & 384 & 317 & 30.12 s & 0/3200 & CPU warm & larger cached set \\
Repeated hotsets & 512 & 369 & 37.13 s & 0/3200 & CPU warm & endpoint in figure \\
\appgroup{7}{Unique-adapter calibration: clean active-window probes}
Unique adapters & 16 & 16 & 1.22 s & 0 & GPU active & exact target \\
Unique adapters & 32 & 32 & 1.34 s & 0 & GPU active & exact target \\
Unique adapters & 64 & 64 & 1.41 s & 0 & GPU active & same-batch frontier \\
\appgroup{7}{Unique-adapter ladder: pressure on local cache}
Unique adapters & 128 & 127 & 5.70 s & 0 & CPU warm & loading begins \\
Unique adapters & 192 & 183 & 10.57 s & 0 & CPU warm & mirrors hotset count \\
Unique adapters & 256 & 236 & 17.79 s & 0 & CPU warm & cache growth \\
Unique adapters & 384 & 317 & 38.37 s & 0 & CPU warm & tail appears \\
Unique adapters & 512 & 369 & 45.36 s & 0 & CPU warm & warm cache still grows \\
Unique adapters & 640 & 410 & 59.01 s & 0 & CPU warm & diminishing cache growth \\
Unique adapters & 768 & 440 & 62.85 s & 0 & CPU warm & near plateau \\
Unique adapters & 896 & 466 & 64.66 s & 0 & CPU warm & near plateau \\
Unique adapters & 1024 & 486 & 66.92 s & 0 & CPU warm & mid-ladder endpoint \\
\appgroup{7}{Unique-adapter high ladder: cache plateau}
Unique adapters & 1280 & 513 & 60.91 s & 0 & CPU warm & plateau persists \\
Unique adapters & 1536 & 525 & 60.95 s & 0 & CPU warm & slow cache growth \\
Unique adapters & 1792 & 537 & 59.95 s & 0 & CPU warm & stable p95 band \\
Unique adapters & 2048 & 550 & 63.14 s & 0 & CPU warm & largest measured target \\
\bottomrule
\end{tabular}%
}
\end{table}

\paragraph{Mixed online-length stress traffic.}
\Cref{tab:app_business_traffic} intentionally leaves the clean warm-path setting. These rows combine mixed output lengths, high concurrency, weak locality, prewarm changes, and different slot limits to show where simple routing choices stop being enough. The sticky-hash row is a negative probe: fixed hashing alone fails to provide cache-aware routing under this stress shape. The GPU-slots-64 row is a bounded probe; it verifies a slot-limit setting over two rounds, with long-run stability left outside this claim.

\begin{table}[H]
\centering
\scriptsize
\setlength{\tabcolsep}{2.8pt}
\renewcommand{\arraystretch}{1.05}
\caption{Mixed online-length traffic on a 2048-adapter catalog. The GPU/CPU column gives the GPU-batch LoRA slot limit and the CPU-cache LoRA limit.}
\label{tab:app_business_traffic}
\fittowidth{%
\begin{tabular}{@{}M{0.18\textwidth}cM{0.10\textwidth}cM{0.10\textwidth}ccccM{0.16\textwidth}@{}}
\toprule
\apphead Probe & C & GPU/CPU & Prewarm & Success & Err. & p50 & p95 & p99 & Interpretation \\
\midrule
\appgroup{10}{Single-route stress envelope}
Single route & 8 & 32/1024 & 128 & 256/256 & 0.00\% & 28.79 & 66.23 & 70.47 & low-concurrency baseline \\
Single route & 16 & 32/1024 & 128 & 499/512 & 2.54\% & 63.18 & 134.03 & 179.74 & tail begins to rise \\
Single route & 32 & 32/1024 & 128 & 969/1024 & 5.37\% & 124.35 & 320.82 & 358.71 & queueing and churn visible \\
Single route & 128 & 32/1024 & 0 & 4032/4096 & 1.56\% & 508.92 & 1028.08 & 1403.44 & high-concurrency stress \\
Single route & 230 & 32/1024 & 0 & 7037/7360 & 4.39\% & 967.10 & 1469.37 & 1717.43 & no-prewarm reference \\
Single route & 230 & 32/2048 & 164 & 6707/7360 & 8.87\% & 926.94 & 1441.86 & 1710.01 & slots alone leave tail high \\
\appgroup{10}{Routing and slot probes}
Sticky hash, 2 replicas & 230 & 32/2048 & 164 & 3737/7360 & 49.23\% & 925.62 & 1435.62 & 1746.58 & naive fixed hashing unstable \\
GPU slots 64, 2 rounds & 230 & 64/2048 & 164 & 460/460 & 0.00\% & 848.98 & 1351.34 & 1690.24 & bounded slot-limit probe \\
\bottomrule
\end{tabular}%
}
\end{table}

\paragraph{Cold-load accounting.}
\Cref{tab:app_cold_load_control} decomposes the cold path into API queueing, shared loads for repeated cache-miss requests, unique-adapter loading, and bounded backpressure. The probes show that the bottleneck is unique cold adapter loading into one engine. Requests wait before generation on LoRA loading; the measured API queue wait is small. Deduplicating identical missing adapters avoids repeated load work, while distinct cold adapters remain separate load jobs. This supports the main-text claim that cold loading is scheduled service work before ordinary sampling latency.

\begin{table}[H]
\centering
\scriptsize
\setlength{\tabcolsep}{3pt}
\renewcommand{\arraystretch}{1.08}
\caption{Cold-load accounting and service protection.}
\label{tab:app_cold_load_control}
\fittowidth{%
\begin{tabular}{@{}M{0.09\textwidth}M{0.22\textwidth}M{0.29\textwidth}M{0.19\textwidth}M{0.17\textwidth}@{}}
\toprule
\apphead ID & Probe & Measured result & Mechanism isolated & Design response \\
\midrule
\appkey{A1} & Same-server no-warm, $N=64$ & Unique-no-warm p95 125.35 s; load median 60.27 s; queue wait median 0.010 s & LoRA loading drives waiting before generation; API queueing is small. & Separate cold loading from sampling latency. \\
\addlinespace
\appkey{A2} & Same adapter vs unique adapters, $N=16$ & Same-adapter load median 1.56 s; unique adapters form 1.47, 2.84, 4.23, $\ldots$, 23.83 s staircase & Repeated cache-miss requests can share one load. & Reduce unique cold misses through routing and prewarming. \\
\addlinespace
\appkey{A3} & Structured add timing, $N=16$ & Engine time median 1.370 s; lock-wait median 10.890 s; total load p95 21.900 s & Exclusive writer/load path serializes unique loads. & Budget unique cold-load capacity per engine. \\
\addlinespace
\appkey{A4} & Bounded cold-load probe & Max in-flight 1 and queue depth 1: a 4-request cold burst loads 2 and rejects 2 & Cold pressure can be observable and retryable. & Prefer bounded backpressure over unbounded blocking. \\
\bottomrule
\end{tabular}%
}
\vspace{-10pt}
\end{table}

\paragraph{Fleet-level active-wave model.}
\Cref{tab:app_fleet_model} uses measured single-engine limits to model a larger MinT deployment under a 2300-distinct-adapter active-wave assumption. The artifact-backed million-scale evidence is the 1M build and audit in \cref{tab:app_1m_catalog_audit}; this model instead asks how many engines would be needed if a large catalog produced that active wave before routing or prewarming restored locality. The 60 s service-level objective is the warm-response target used for the throughput floor, and 2.57 req/s/engine is the observed warm-throughput input for that row. The cold-load rows size the separate case in which many selected adapters are cold at first touch.

\begin{table}[H]
\centering
\scriptsize
\setlength{\tabcolsep}{3.4pt}
\renewcommand{\arraystretch}{1.08}
\caption{Legacy capacity-planning sketch from measured per-engine limits to a 2300-distinct-adapter active-wave stress envelope. The 1M catalog itself is now measured in \cref{tab:app_1m_catalog_audit}; this table is retained only as fleet-sizing intuition.}
\label{tab:app_fleet_model}
\fittowidth{%
\begin{tabular}{@{}M{0.22\textwidth}M{0.25\textwidth}ccM{0.23\textwidth}@{}}
\toprule
\apphead Axis & Sizing rule & Engines & GPUs & System interpretation \\
\midrule
\appgroup{5}{Warm-path placement}
Warm distinct concurrency & $\lceil 2300/64 \rceil$ & 36 & 144 & ideal placement by same-batch adapter diversity \\
Warm headroom band & $36\times\{1.2,1.33,1.5\}$ & 44--54 & 176--216 & margin for skew, retries, and imperfect routing \\
Warm throughput floor & 60 s SLO; 2.57 req/s/engine & 15 & 60 & throughput-only floor; ignores distinct placement \\
\appgroup{5}{Cold-path isolation}
Cold-load rate & 38.3 cold LoRA/s; 0.7/engine & 55 & 220 & rates the cold-load service with burst tail tracked separately \\
Cold-burst isolation & 2300 cold uniques; $\le 32$/engine & 72 & 288 & isolates worst-case first touch through prewarm or backpressure \\
\bottomrule
\end{tabular}%
}
\end{table}

\end{appendices}

\end{document}